\documentclass[twoside]{article}

\usepackage[accepted]{aistats2015}
\usepackage{amsmath}
\usepackage{bm}
\usepackage{dsfont}
\usepackage{graphicx}
\usepackage{multicol}
\usepackage[authoryear]{natbib}
\usepackage{tikz}
\usepackage{times}
\usepackage{subfigure} 
\usepackage{xspace}
\usepackage{xcolor}

\usetikzlibrary{bayesnet,matrix}

\newcommand{\METHOD}{Consensus Message Passing\@\xspace}
\newcommand{\Method}{Consensus message passing\@\xspace}
\newcommand{\method}{consensus message passing\@\xspace}
\newcommand{\MTD}{CMP\@\xspace}

\newcommand{\eg}{\textit{e.g.}\@\xspace}
\newcommand{\ie}{\textit{i.e.}\@\xspace}

\newcommand{\etal}{\textit{et~al.}\@\xspace}

\newcommand{\mycaption}[2]{\caption{\textbf{#1.}\xspace#2}}

\newcommand{\argmax}{\operatornamewithlimits{argmax}}

\makeatletter
\renewcommand{\thesubfigure}{\alph{subfigure}}
\renewcommand{\@thesubfigure}{(\thesubfigure)\hskip\subfiglabelskip}
\makeatother

\tikzset{
  cluster/.style={rectangle, minimum height=0.4cm, minimum width=0.4cm, inner sep=0.05cm, draw},
  mylatent/.style={circle, minimum height=0.5cm, minimum width=0.5cm, inner sep=0.05cm, draw},
  myfactor/.style={rectangle, minimum height=0.05cm, minimum width=0.05cm, fill=black, scale=0.75},
  myinitfactor/.style={rectangle, minimum height=0.1cm, minimum width=0.1cm, fill=red, scale=0.75},
}

\begin{document}

%
\runningtitle{Consensus Message Passing for Layered Graphical Models}

%

\renewcommand*{\thefootnote}{\fnsymbol{footnote}}

\twocolumn[

\aistatstitle{Consensus Message Passing for Layered Graphical Models}

\aistatsauthor{ Varun Jampani\footnotemark[2] \And S. M. Ali Eslami\footnotemark[2], Daniel Tarlow, Pushmeet Kohli and John Winn }

\aistatsaddress{ MPI for Intelligent Systems, T\"{u}bingen \And Microsoft Research, Cambridge } ]

\footnotetext[2]{The first two authors contribute equally to this work.}


\begin{abstract}
Generative models provide a powerful framework for probabilistic reasoning. However, in many domains their use has been hampered by the practical difficulties of inference. This is particularly the case in computer vision, where models of the imaging process tend to be large, loopy and layered. For this reason bottom-up conditional models have traditionally dominated in such domains. We find that widely-used, general-purpose message passing inference algorithms such as Expectation Propagation (EP) and Variational Message Passing (VMP) fail on the simplest of vision models. With these models in mind, we introduce a modification to message passing that learns to exploit their layered structure by passing \textit{consensus} messages that guide inference towards good solutions. Experiments on a variety of problems show that the proposed technique leads to significantly more accurate inference results, not only when compared to standard EP and VMP, but also when compared to competitive bottom-up conditional models.
\end{abstract}

\vspace{-0.2cm}
\section{Introduction}
\label{sec:introduction}

Generative models provide a powerful framework for probabilistic reasoning and are applicable across a wide variety of domains, including computational biology, natural language processing, and computer vision. For example, in computer vision, one can use graphical models to express the process by which a face is lit and rendered into an image, incorporating knowledge of surface normals, lighting and even the approximate symmetry of human faces. Models that make effective use of this information will generalize well, and they will require less labelled training data than their unstructured counterparts (\eg random forests or neural networks) in order to make accurate predictions.

Perhaps the most significant challenge of the generative modelling framework is that inference can be very hard. Sampling-based methods run the risk of slow mixing, while message passing-based methods (which are the focus of this work) can converge slowly, converge to bad solutions, or fail to converge at all. Whilst significant efforts have been made to improve the accuracy of message passing algorithms (\eg by using structured variational approximations), many challenges remain, including difficulty of implementation, the problem of computational cost and the question of how the structured approximation should be chosen. The present work aims to alleviate these problems for general-purpose message-passing algorithms.

Our starting observation was that general purpose message passing inference algorithms (\eg EP and VMP; \citealp{Minka2001,Winn2005}) fail on even the simplest of computer vision models. We claim that in these models the failure can be attributed to the algorithms' inability to determine the values of a relatively small number of influential variables which we call `global' variables. Without accurate estimation of these global variables, it can be very difficult for message passing to make meaningful progress on the other variables in the model.

Latent variables in vision models are often organised in a layered structure, where the observed image pixels are at the bottom and high-level scene parameters are at the top. Additionally, knowledge about the values of the variables at level $l$ is sufficient to reason about any global variable at layer $l+1$. With these properties in mind, we develop a method called \emph{\METHOD} (\MTD) that learns to exploit such layered structures and estimate global variables during the early stages of inference. 

Experimental results on a variety of problems show that \MTD leads to significantly more accurate inference results whilst preserving the computational efficiency of standard message passing. The implication of this work is twofold. First, it adds a useful tool to the toolbox of techniques for improving general-purpose inference, and second, in doing so it overcomes a bottleneck that has restricted the use of model-based machine learning in computer vision.

\vspace{-0.2cm}
\section{\METHOD}
\label{sec:method}

\Method exploits the layered characteristic of vision models in order to overcome the aforementioned inference challenges. For illustration, two layers of latent variables of such a model are shown in Fig.~\ref{fig:types-a} using factor graph notation (black). Here the latent variables below ($\mathbf{h}^b = \{ h^b_k\}$) are a function of the latent variables above ($\mathbf{h}^a = \{ h^a_k\}$) and the global variables $x$ and $y$ (where $k$ ranges over pixels; in this case $|k|=3$). As we will see in the experiments that follow, this is a recurring pattern that appears in many models of interest in vision. For example, in the case of face modeling, the $\mathbf{h}^a$ variables correspond to the normals $\mathbf{n}_i$, the global variable $x$ to the light vector $\mathbf{l}$, and $\mathbf{h}^b$ to the shading intensities $s_i$ (see Fig.~\ref{fig:shading-model}).

Our reasoning follows a recursive structure. Assume for a moment that in Fig.~\ref{fig:types-a}, the messages from the layer below to the inter-layer factors (blue) are both informative and accurate (\eg due to being close to the observed pixels). We will refer to these messages collectively as \textit{contextual messages}. It would be desirable, for purposes of both speed and accuracy, that we could ensure that the messages sent to the layer above ($\mathbf{h}^a$) also possess the same properties. If we had access to an oracle that could give us the correct belief for the global variables ($x$ and $y$) for the image, we could send accurate initial messages from $x$ and $y$ and in one step compute informative and accurate messages from the inter-layer factors to the layer above.

In practice, however, we do not have access to such an oracle. In this work we train regressors to \textit{predict} the values of the global variables given all the messages from the layer below. Should this prediction be good enough, the messages to the layer above will be informative and accurate, and the inductive argument will hold. We describe how these regressors are trained in Sec.~\ref{sec:training}. To summarize, the approach consists of the following two components:

\begin{enumerate}

\item Before inference, for each global variable in different layers of the model, we train a regressor to predict some oracle's value for the target variable given the values of all the messages from the layer below (\ie the \textit{contextual messages}, Fig.~\ref{fig:types-a}, blue),

\item During inference, each regressor sends this belief in the form of a \textit{consensus message} (Fig.~\ref{fig:types-a}, red) to its target variable.

\end{enumerate}

In some models it will be useful to employ a second type of \MTD, displayed graphically in Fig.~\ref{fig:types-b}, where global layer variables are absent and loops in the graphical model are due to global variables in other layers. Here, a consensus message is sent to each variable in the latent layer above, given all the contextual messages.

\begin{figure}[t]
	\centering
	\subfigure[Type A]{
		\begin{tikzpicture}
			\matrix at (0, 0.2) [matrix, column sep=0.1cm, row sep=0.21cm,ampersand replacement=\&]
			{
				\node { }; \&
				\node (y1e) { $\vdots$ }; \&
				\node (y2e) { $\vdots$ }; \&
				\node (y3e) { $\vdots$ }; \&
				\node { }; \\

				\node (a) [mylatent] { $x$ }; \&
				\node (y1) [mylatent] { $h^a_1$ }; \&
				\node (y2) [mylatent] { $h^a_2$ }; \&
				\node (y3) [mylatent] { $h^a_3$ }; \&
				\node (b) [mylatent] { $y$ }; \\

				\& \& \& \& \\

				\node { }; \&
				\node (cl1) [myfactor] {  }; \&
				\node (cl2) [myfactor] {  }; \&
				\node (cl3) [myfactor] {  }; \&
				\node { }; \& \\

				\& \& \& \& \\
				\& \& \& \& \\
				\& \& \& \& \\

				\node { }; \&
				\node (x1) [mylatent] { $h^b_1$ }; \&
				\node (x2) [mylatent] { $h^b_2$ }; \&
				\node (x3) [mylatent] { $h^b_3$ }; \&
				\node { }; \\

				\node { }; \&
				\node (x1e) { $\vdots$ }; \&
				\node (x2e) { $\vdots$ }; \&
				\node (x3e) { $\vdots$ }; \&
				\node { }; \\	
			};

			\draw[red!20] (-1.9, 0) -- (1.8, 0);
			\draw[red, dotted] (-1.9, 0) -- (1.8, 0);

			\fill (a.south) ++ (0, -1.02) circle (3pt) [fill=red] { };
			\fill (b.south) ++ (0, -1.02) circle (3pt) [fill=red] { };

			\node[yshift=-1.35cm] at (a.south) { $\textcolor{red}{\Delta^x}$ };
			\node[yshift=-1.35cm] at (b.south) { $\textcolor{red}{\Delta^y}$ };

			\draw[-stealth, cyan] (0.1, -0.2) -- (0.1, 0.2);
			\draw[-stealth, cyan] (0.87, -0.2) -- (0.87, 0.2);
			\draw[-stealth, cyan] (-0.7, -0.2) -- (-0.7, 0.2);

			\fill (0.1, 0) circle (1pt) [fill=cyan] { };
			\fill (0.87, 0) circle (1pt) [fill=cyan] { };
			\fill (-0.7, 0) circle (1pt) [fill=cyan] { };

			\draw[-stealth, red] (1.41, 0.25) -- (1.41, 0.9);
			\draw[-stealth, red] (-1.52, 0.25) -- (-1.52, 0.9);

			\draw (y1.north) -- (y1e);
			\draw (y2.north) -- (y2e);
			\draw (y3.north) -- (y3e);

			\draw (a.south) -- (cl1.north);
			\draw (a.south) -- (cl2.north);
			\draw (a.south) -- (cl3.north);
			\draw (b.south) -- (cl1.north);
			\draw (b.south) -- (cl2.north);
			\draw (b.south) -- (cl3.north);
			\draw (y1.south) -- (cl1);
			\draw (y2.south) -- (cl2);
			\draw (y3.south) -- (cl3);
			\draw [->] (cl1) -- (x1.north);
			\draw [->] (cl2) -- (x2.north);
			\draw [->] (cl3) -- (x3.north);

			\draw (x1.south) -- (x1e);
			\draw (x2.south) -- (x2e);
			\draw (x3.south) -- (x3e);

		\end{tikzpicture}
		\label{fig:types-a}
	}
	\hfill
	\subfigure[Type B]{
		\begin{tikzpicture}

			\matrix at (0, 0.2) [matrix, column sep=0.4cm, row sep=0.217cm,ampersand replacement=\&]
			{
				\node (y1e) { $\vdots$ }; \&
				\node (y2e) { $\vdots$ }; \&
				\node (y3e) { $\vdots$ }; \\

				\node (y1) [mylatent] { $h^a_1$ }; \&
				\node (y2) [mylatent] { $h^a_2$ }; \&
				\node (y3) [mylatent] { $h^a_3$ }; \\

				\& \& \\

				\node (cl1) [myfactor] {  }; \&
				\node (cl2) [myfactor] {  }; \&
				\node (cl3) [myfactor] {  }; \\

				\& \& \\
				\& \& \\
				\& \& \\

				\node (x1) [mylatent] { $h^b_1$ }; \&
				\node (x2) [mylatent] { $h^b_2$ }; \&
				\node (x3) [mylatent] { $h^b_3$ }; \\

				\node (x1e) { $\vdots$ }; \&
				\node (x2e) { $\vdots$ }; \&
				\node (x3e) { $\vdots$ }; \\	
			};

			\draw[red!20] (-1.9, 0) -- (1.6, 0);
			\draw[red, dotted] (-1.9, 0) -- (1.6, 0);

			\fill (y1.south) ++ (-0.5, -0.95) circle (3pt) [fill=red] { };
			\fill (y2.south) ++ (-0.5, -0.95) circle (3pt) [fill=red] { };
			\fill (y3.south) ++ (-0.5, -0.95) circle (3pt) [fill=red] { };

			\node[xshift=-0.5cm, yshift=-1.25cm] at (y1.south) { $\textcolor{red}{\Delta^1}$ };
			\node[xshift=-0.5cm, yshift=-1.25cm] at (y2.south) { $\textcolor{red}{\Delta^2}$ };
			\node[xshift=-0.5cm, yshift=-1.25cm] at (y3.south) { $\textcolor{red}{\Delta^3}$ };

			\draw[-stealth, cyan] (0.17, -0.2) -- (0.17, 0.2);
			\draw[-stealth, cyan] (1.27, -0.2) -- (1.27, 0.2);
			\draw[-stealth, cyan] (-0.92, -0.2) -- (-0.92, 0.2);

			\fill (0.17, 0) circle (1pt) [fill=cyan] { };
			\fill (1.27, 0) circle (1pt) [fill=cyan] { };
			\fill (-0.92, 0) circle (1pt) [fill=cyan] { };

			\draw[-stealth, red] (-0.46, 0.23) -- (-0.13, 0.87);
			\draw[-stealth, red] (-1.53, 0.23) -- (-1.2, 0.87);
			\draw[-stealth, red] (0.61, 0.23) -- (0.94, 0.87);

			\draw (y1.north) -- (y1e);
			\draw (y2.north) -- (y2e);
			\draw (y3.north) -- (y3e);

			\draw (y1.south) -- (cl1);
			\draw (y2.south) -- (cl2);
			\draw (y3.south) -- (cl3);
			\draw [->] (cl1) -- (x1.north);
			\draw [->] (cl2) -- (x2.north);
			\draw [->] (cl3) -- (x3.north);

			\draw (x1.south) -- (x1e);
			\draw (x2.south) -- (x2e);
			\draw (x3.south) -- (x3e);

		\end{tikzpicture}
		\label{fig:types-b}
	}
	\mycaption{\Method}{Vision models tend to be large, layered and loopy. (a)~Two adjacent layers of the latent variables of a model of this kind (black). In \MTD, consensus messages (red) are computed from contextual messages (blue) and sent to global variables ($x$ and $y$), guiding inference in the layer. (b)~\Method of a different kind for situations where loops in the graphical model are due to global variables in other layers.}
	\label{fig:types}
\end{figure} 

Any message passing schedule can be used subject to the constraint that the consensus messages are given maximum priority within a layer and that they are sent bottom up. Naturally, a consensus message can only be sent once its contextual messages have been computed. It is desirable to be able to ensure that the fixed point reached under this scheme is also a fixed point of standard message passing in the model. One approach for this is to reduce the certainty of the consensus messages over the course of inference, or to only pass them in the first few iterations. In our experiments we found that even passing consensus messages only in the first iteration led to accurate inference, and therefore we follow this strategy for the remainder of the paper. It is worth emphasizing that message-passing equations remain unchanged and we used the same scheduling scheme in all our experiments (\ie no need for manual tuning).

It is important to highlight a crucial difference between \method and heuristic \textit{initialization}. In the latter, predictions are made from the \textit{observations} no matter how high up in the hierarchy the target variable is, whereas in \MTD predictions are made using \textit{messages} that are sent from variables immediately below the target variables of interest.  The CMP prediction task will be much simpler, since the relationship between the target variables and the variables in the layer immediately below is much less complex than the relationship between the target variables and the observations.  Furthermore, we know from the layered structure of the model that all relevant information from the observations is contained in the variables in the layer below.  This is because 
 target variables at layer $l+1$ are conditionally independent of all layers $l-1$ and below, given the values of layer $l$. 

One final note on the capacity of the regressors. Of course it is true that an infinite capacity regressor can make perfect predictions given enough data (whether using \MTD or heuristic initialization). However we are interested in practical ways of obtaining accurate results for models of increasing complexity, where lack of capable regressors and unlimited data is inevitable. One important feature of \MTD is that it makes use of predictors in a scalable way, since regressions are only made between adjacent latent layers.

\section{Predictor Training}
\label{sec:training}

To recap, we wish to perform inference in a layered model of observed variables $\mathbf{X}$ with latent variables $\mathbf{H}$. Each predictor $\Delta^t$ (with target $t$) is a function of a collection of its contextual messages $\mathbf{c} = \{ c_k \}$ (incoming from the latent layer below $\mathbf{h}^b$), that produces the consensus message $m$, \ie $m = \Delta^t(\mathbf{c}).$

We adopt an approach in which we \textit{learn} a function for this task that is parameterized by $\bm{\theta}$, \ie $\overline{m} \equiv f(\mathbf{c}|\bm{\theta}).$ This can be seen as an instance of the canonical regression task. For a given family of regressors $f$, the goal of training is to find parameters $\bm{\theta}$ that capture the relationship between context and consensus message pairs $\{ (\mathbf{c}_d, m_d) \}_{d=1...D}$ in some set of training examples.

\subsection{Choice of predictor training data}

First we discuss how this training data is obtained. There can be at least three different sources:

\textbf{1.\,\,Beliefs at convergence.} This technique is only useful if standard message passing works but is slow. Standard message passing inference is run in the model for a large number of iterations and for a collection of different observations $\{ \mathbf{X}_d \}$. Message passing is scheduled in precisely the same way as it would be if \MTD were present, however no consensus messages are sent. For each observation $\mathbf{X}_d$, the collection of the marginals of the latent variables in the layer below the predictor ($\mathbf{h}^b_d = \{ h^b_{dk} \}$, see \eg Fig.~\ref{fig:types-a}) at the \textit{first} iteration of message passing is considered to be the context $\mathbf{c}_d$, and the marginal of the target variable $t$ at the \textit{last} iteration of message passing is considered to be the oracle message $m_d$. The aim is that during inference on new problems, a predictor trained in this way would send messages that \textit{accelerate} convergence to the fixed-point that message passing would have reached by itself anyway.

\textbf{2.\,\,Samples from the model.} This technique is useful if standard message passing fails to reach good fixed points no matter how long it is run for. First a collection of samples from the model is generated, giving us for each sample both the observation $\mathbf{X}_d$ and its corresponding latent variables $\mathbf{H}_d$. Standard message passing inference is then run on the observations $\{ \mathbf{X}_d \}$ only for a single iteration. Message passing is scheduled as before. For each observation $\mathbf{X}_d$, the marginals of the latent variables in the layer below $\mathbf{h}^b_d$ at the \textit{first} iteration of message passing is the context $\mathbf{c}_d$, and the oracle message $m_d$ is considered to be a point-mass centered at the sampled value of the target variable $t$. The aim is that during inference on new problems, a predictor trained in this way would send messages that guide inference to a fixed-point in which the marginal of the target variable $t$ is close to its sampled value.

\textbf{3.\,\,Labelled data.} As above, except the latent variables of interest $\mathbf{h}_d$ are set from real data instead of being sampled from the model. The oracle message $m_d$ is therefore a point-mass centered at the label provided for the target variable $t$ for observation $\mathbf{X}_d$. The aim is that during inference on new problems, a predictor trained in this way would send messages that guide inference to a fixed-point in which the marginal of the target variable $t$ is close to its labelled value, even in the presence of a degree of model mismatch. We demonstrate each of the strategies in the experiments in Sec.~\ref{sec:experiments}.

\vspace{-0.1cm}
\subsection{Random regression forests}

We wish to learn a mapping $f$ from contextual messages $\mathbf{c}$ to the consensus message $m$ from training data $\{ (\mathbf{c}_d, m_d) \}_{d=1...D}$. This is challenging since the inputs and outputs of the regression problem are messages (\ie distributions), and special care needs to be taken to account for this fact. We follow closely the methodology of~\cite{Eslami2014}, in which random forests are used to predict outgoing EP messages from a factor. A detailed description of our random forest implementation is provided in the supplementary material. For a review of forests see~\cite{Criminisi2013}.

\vspace{-0.2cm}
\section{Experiments}
\label{sec:experiments}

We first illustrate the application of \MTD to two diagnostic models: one of circles and a second of squares. We then use the approach to improve inference in a more challenging vision model: that of intrinsic images of faces. In the first experiment the predictors are trained on beliefs at convergence, in the second on samples from the model, and in the third on annotated labels, showcasing various use-cases of \MTD. We show that in all cases the proposed technique leads to significantly more accurate inference results whilst preserving the computational efficiency of message passing. The experiments were performed in Infer.NET~\citep{InferNET2012} using default settings, unless stated otherwise. We set the number of trees in each forest to 8.

\subsection{A generative model of circles}
\label{sec:circle}

\begin{figure}[t]
	\centering
	\subfigure[]{
		\setlength\fboxsep{-0.3mm}
		\setlength\fboxrule{0.5pt}
		\parbox[b]{3.1cm}{
			\fbox{\includegraphics[width=\linewidth]{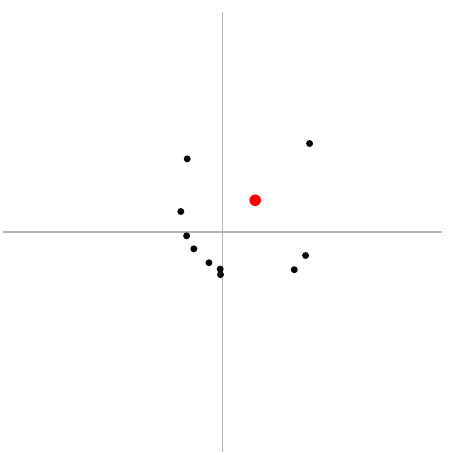}} \\
			\vspace{1mm}
			\fbox{\includegraphics[width=\linewidth]{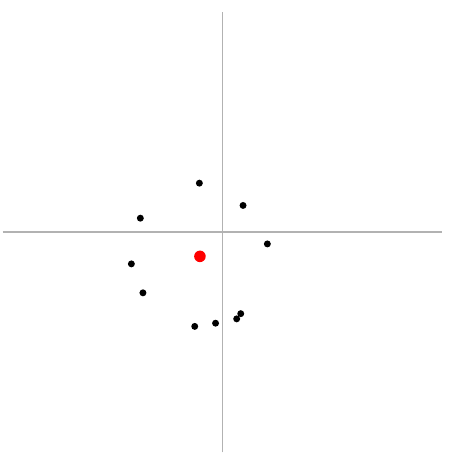}}
		}
		\hspace{0.1cm}
		\label{fig:circle-data}
	}
	\subfigure[]{
		\begin{tikzpicture}

			\draw[red!20] (-2.3, 2.5) -- (2, 2.5);
			\draw[red, dotted] (-2.3, 2.5) -- (2, 2.5);

			\node[obs] 											(x)			{$\mathbf{x}_i$}; %
			\node[latent, above=of x, yshift=-2mm]                			(z)			{$\mathbf{z}_i$}; %

			\factor[above=of x]				 					{noise}		{left:Gaussian} {} {}; %
			\factor[above=of z, yshift=10mm] 					{sum}		{left:Sum} {} {}; %

			\node[latent, right=of sum]                			(c)			{$\mathbf{c}$}; %
			\node[latent, above=of sum, yshift=-7mm]   			(p)			{$\mathbf{p}_i$}; %

			\factor[above=of p] 								{circle}	{above:Circle\,\,\,\,\,} {} {}; %
			\factor[above=of c] 								{pc}		{} {} {}; %

			\node[latent, left=of circle]                		(a)			{$a_i$}; %
			\node[latent, right=of circle]                		(r)			{$r$}; %

			\factor[above=of a] 								{pa}		{} {} {}; %
			\factor[above=of r] 								{pr}		{} {} {}; %

			\factoredge {z} 		{noise} 		{x}; %
			\factoredge {p} 		{sum} 			{}; %
			\factoredge {a} 		{circle} 		{p}; %
			\factoredge {r} 		{circle} 		{}; %
			\factoredge {c} 		{sum} 			{z}; %
			\factoredge {} 			{pc} 			{c}; %
			\factoredge {} 			{pa} 			{a}; %
			\factoredge {} 			{pr} 			{r}; %

			\plate {} {(pa) (a) (p) (z) (x)} {}; %

			\fill (c.south) ++ (0, -0.52) circle (3pt) [fill=red] { };

			\draw [-stealth, red] (1.45, 2.65) -- (1.45, 2.95);

			\node[yshift=-0.9cm] at (c.south) { $\textcolor{red}{\Delta^c}$ };)

		\end{tikzpicture}
		\label{fig:circle-model}
	}
	\mycaption{The circle problem}{(a)~Given a sample of points on a circle (black), we wish to infer the circle's center (red) and its radius. Two sets of samples are shown. (b)~The graphical model for this problem.}
	\label{fig:circle}
\end{figure}

\begin{figure}[t]
	\centering
	\subfigure[Center]{
		\includegraphics[width=0.46\linewidth]{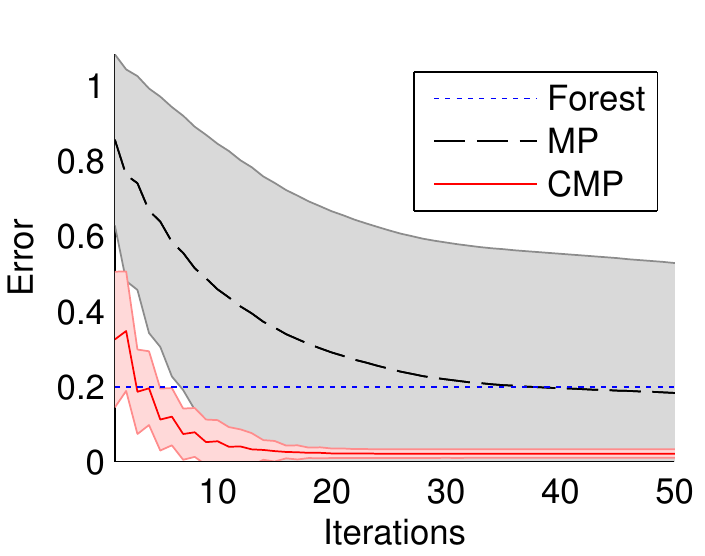}
	}
	\subfigure[Radius]{
		\includegraphics[width=0.46\linewidth]{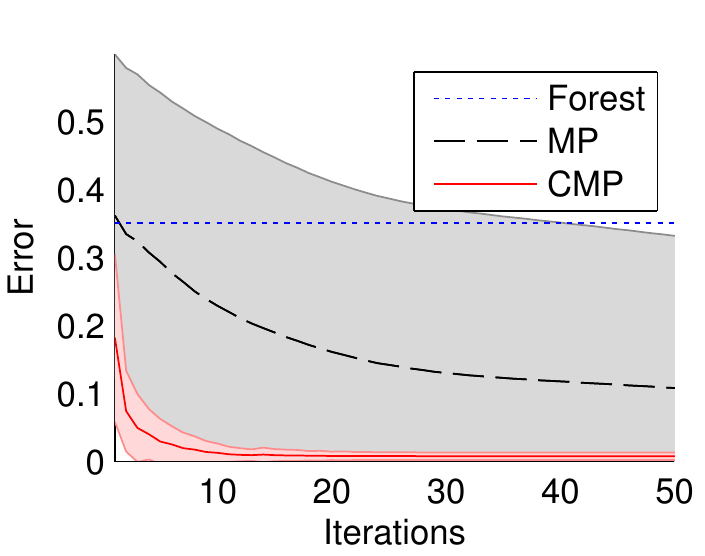}
	}
	\mycaption{Accelerated inference using \MTD}{(a)~Distance of the mean of the marginal posterior of $c$ from its true value as a function of number of inference iterations (Forest: direct prediction, MP: standard VMP, \MTD: VMP with consensus). \Method significantly accelerates convergence. (b)~Similar plot for radius $r$. }
	\label{fig:circle-results}
\end{figure}

We begin by studying the behaviour of standard message passing on a simplified Gauss and Ceres problem~\citep{Teets1999}. We use this example to highlight the fact that although inference may require many iterations of message passing, message initialization can have a significant effect on the speed of convergence, and to demonstrate how this can be done automatically using \MTD.

Given a noisy sample of points $\mathbf{X} = \{ \mathbf{x}_i \}_{i=1...N}$ on a circle in the 2D plane (Fig.~\ref{fig:circle-data}, black, $\mathcal{N}(0,0.01)$ noise on each axis), we wish to infer the coordinates of the circle's center $\mathbf{c}$ (Fig.~\ref{fig:circle-data}, red) and its radius $r$. We can express the data generation process using a graphical model (Fig.~\ref{fig:circle-model}). The Cartesian point $(0, r)$ is rotated $a_i$ radians to generate $\mathbf{p}_i$, then translated by $\mathbf{c}$ to generate the latent $\mathbf{z}_i$, which finally produces the noisy observation $\mathbf{x}_i$.  This model can be expressed in a few lines of code in Infer.NET. The circle model is interesting for our purposes since it is both layered (the $\mathbf{z}_i$s, $\mathbf{p}_i$s and $a_i$s each form a layer) and loopy (due to the presence of two variables outside the plate).

Vanilla message passing inference in this model can take a surprisingly large number of iterations to converge. We draw 10 points $\{ \mathbf{x}_i \}$ from circles with random centers and radii, run VMP and record the accuracy of the marginals of the latent variables at each iteration. We repeat the experiment 50 times and plot results in Fig.~\ref{fig:circle-results} (dashed black). As can be seen from the figure, the marginals contain significant errors even after 50 iterations of message passing.

We then experiment with \method. A predictor $\Delta^c$ is trained to send a consensus message to $\mathbf{c}$ in the initial stages of inference, given the messages coming up from all of the $\mathbf{z}_i$ (indicated graphically in Fig.~\ref{fig:circle-model}, red). The predictor is trained on final beliefs at 100 iterations of standard message passing on $D=500$ sample problems.

As can be seen in Fig.~\ref{fig:circle-results} (red), this single consensus message has the effect of significantly increasing the rate of convergence (as indicated by slope) and also inference robustness (as indicated by error bars). For comparison we also plot how well a regressor of the same capacity as the one used by \MTD can directly estimate the latent variables without using the graphical model in Fig.~\ref{fig:circle-results} (blue). \Method gives us the best of both worlds in this example: speed that is more comparable to one-shot bottom-up prediction and the accuracy of message passing inference in a good model for the problem.

\begin{figure}[t]
	\centering
	\subfigure[]{
		\setlength\fboxsep{-0.2mm}
		\setlength\fboxrule{0.7pt}
		\parbox[b]{2.3cm}{
			\fbox{\includegraphics[width=0.45\linewidth]{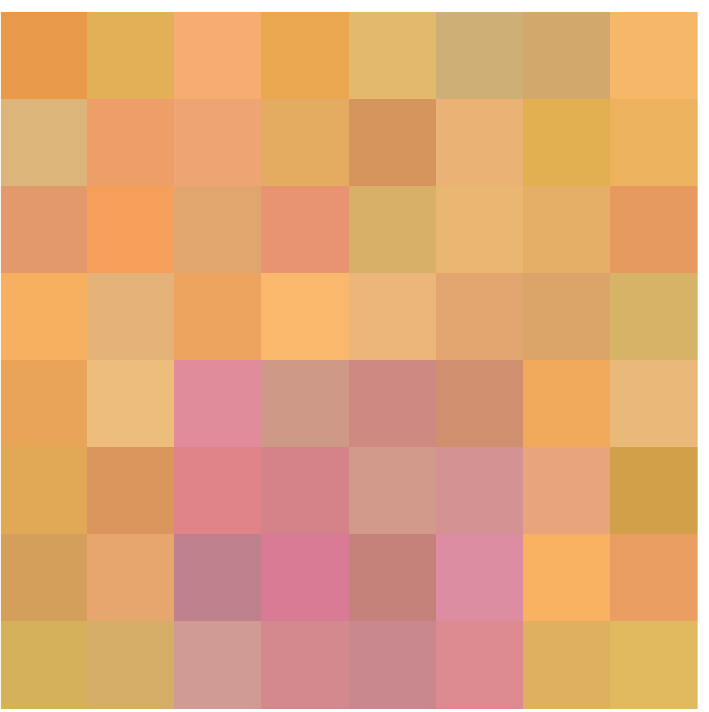}} \hspace*{0mm}
			\fbox{\includegraphics[width=0.45\linewidth]{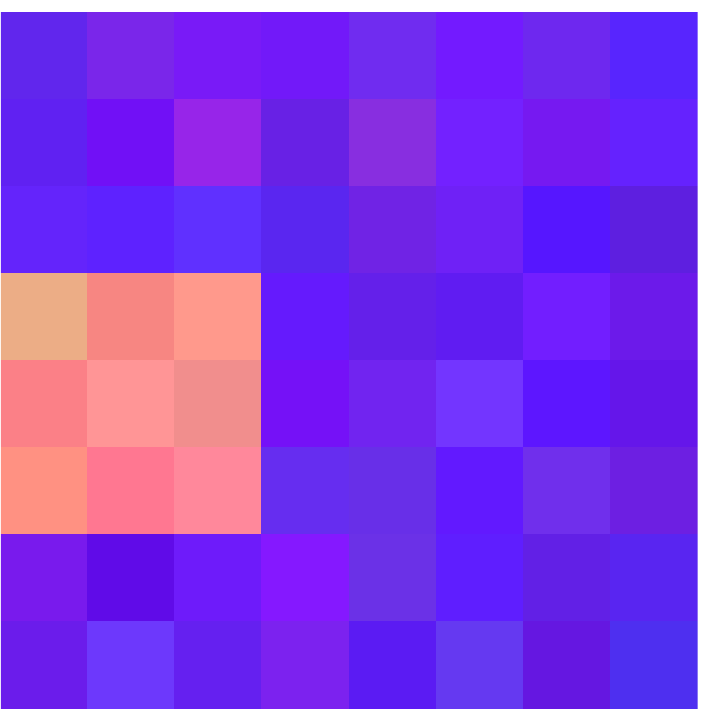}} \vspace*{-2.4mm} \\
			\fbox{\includegraphics[width=0.45\linewidth]{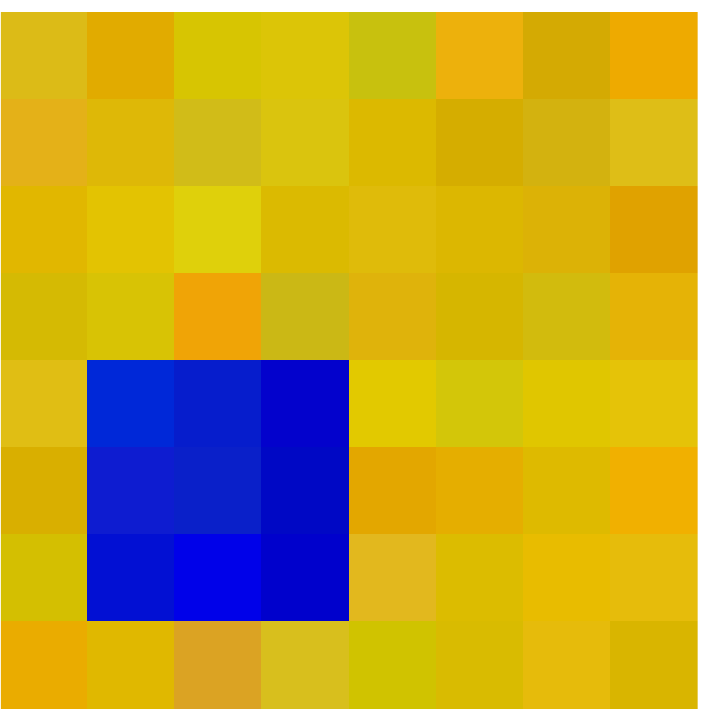}} \hspace*{0mm}
			\fbox{\includegraphics[width=0.45\linewidth]{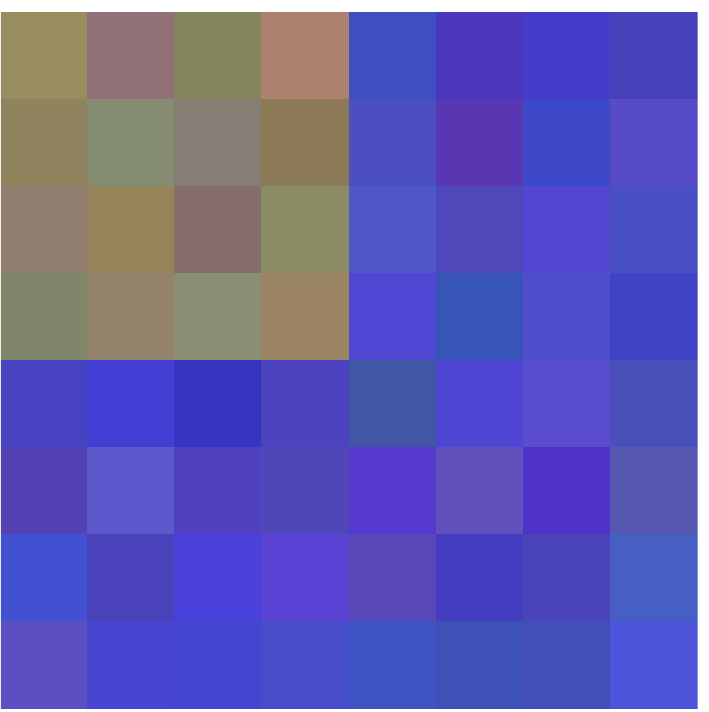}} \vspace*{-2.4mm} \\
			\fbox{\includegraphics[width=0.45\linewidth]{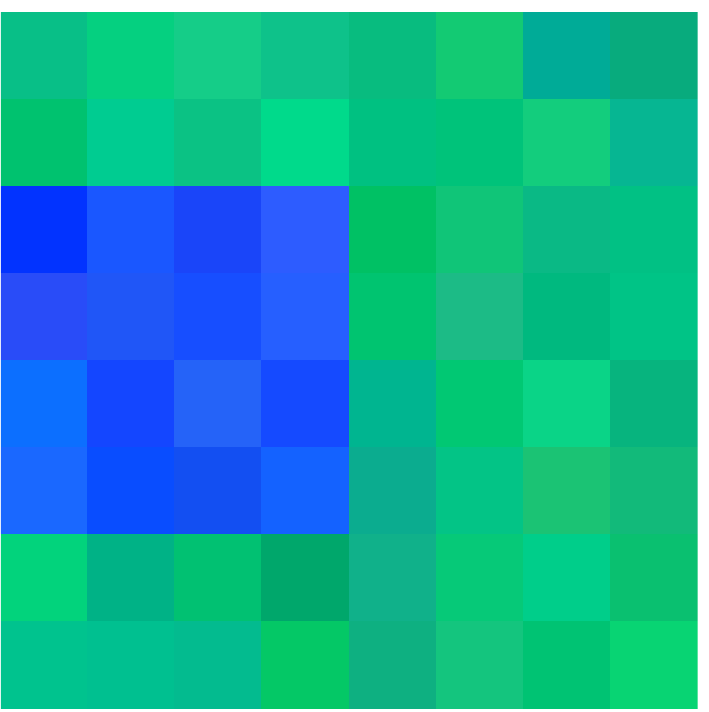}} \hspace*{0mm}
			\fbox{\includegraphics[width=0.45\linewidth]{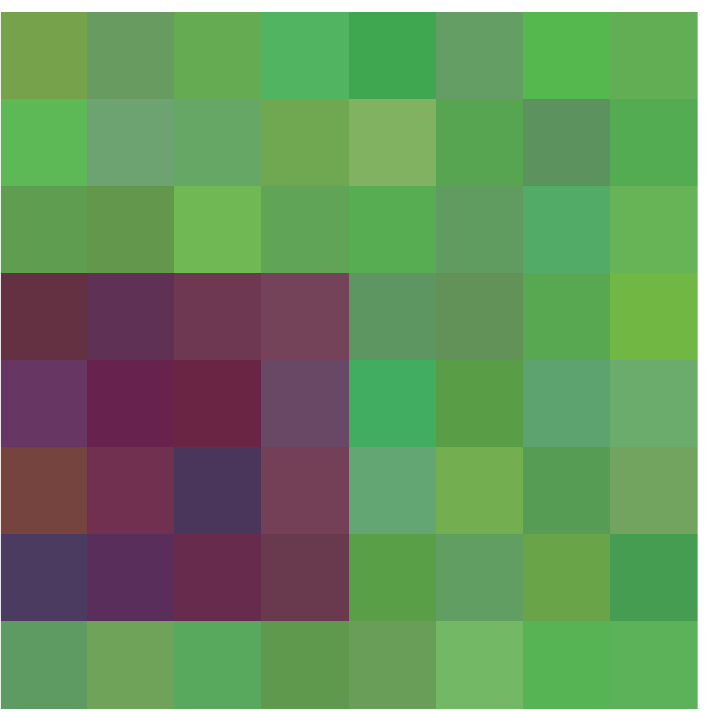}} \vspace*{-2.4mm} \\
			\fbox{\includegraphics[width=0.45\linewidth]{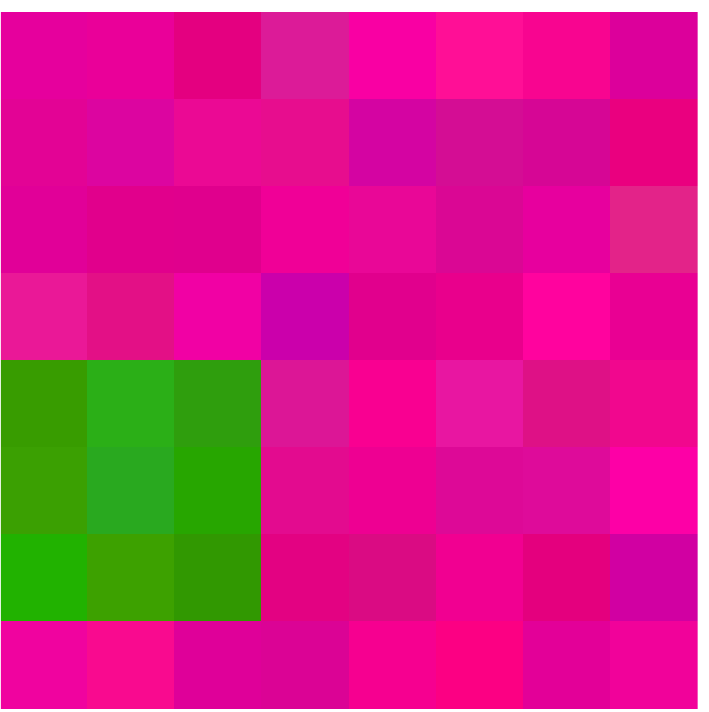}} \hspace*{0mm}
			\fbox{\includegraphics[width=0.45\linewidth]{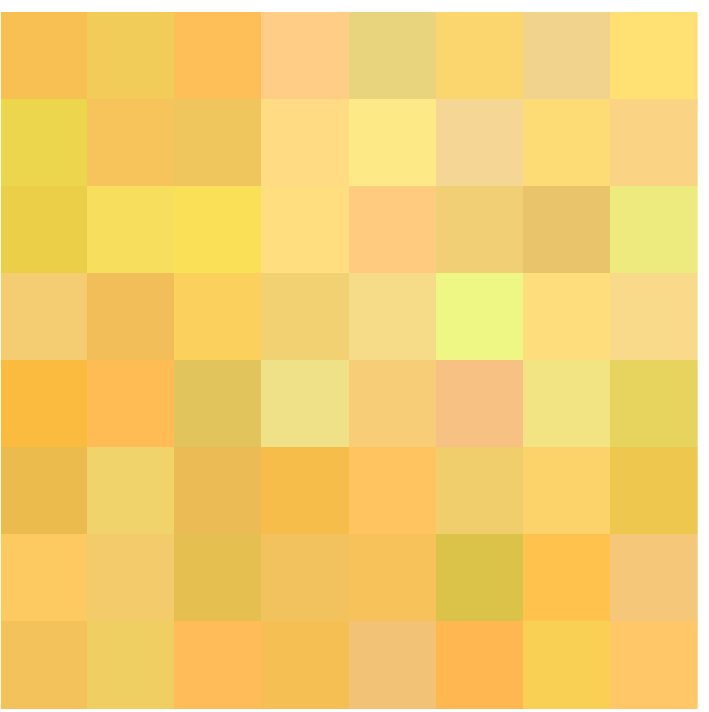}} \vspace*{-2.4mm} \\
			\fbox{\includegraphics[width=0.45\linewidth]{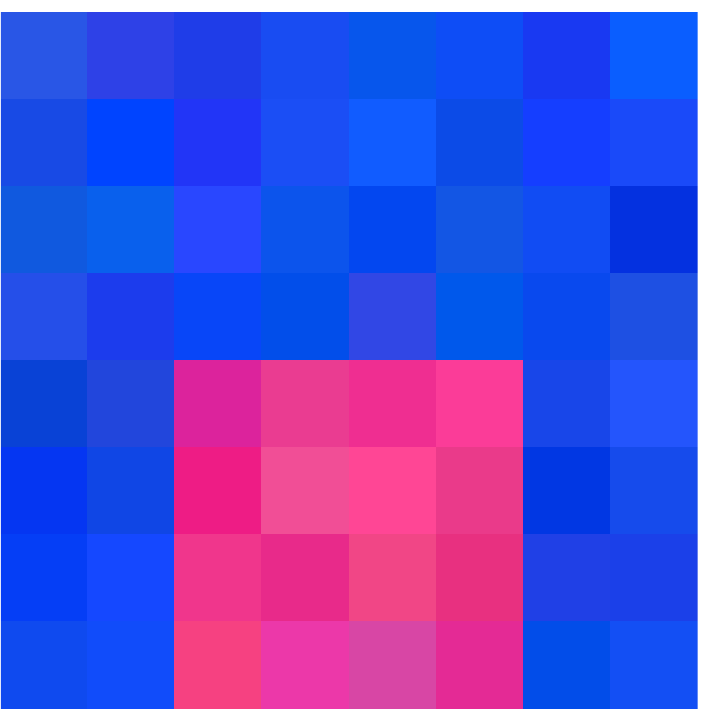}} \hspace*{0mm}
			\fbox{\includegraphics[width=0.45\linewidth]{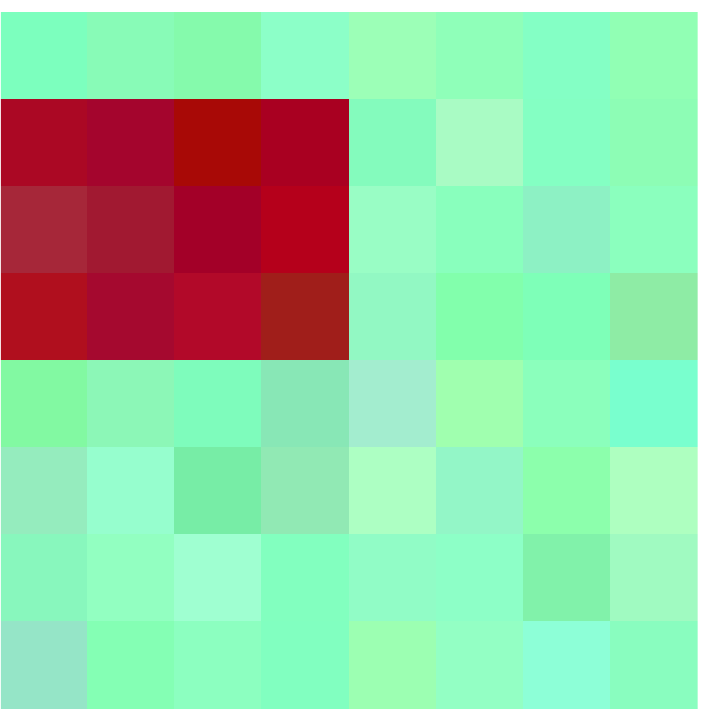}} \vspace*{-2.4mm} \\
			\fbox{\includegraphics[width=0.45\linewidth]{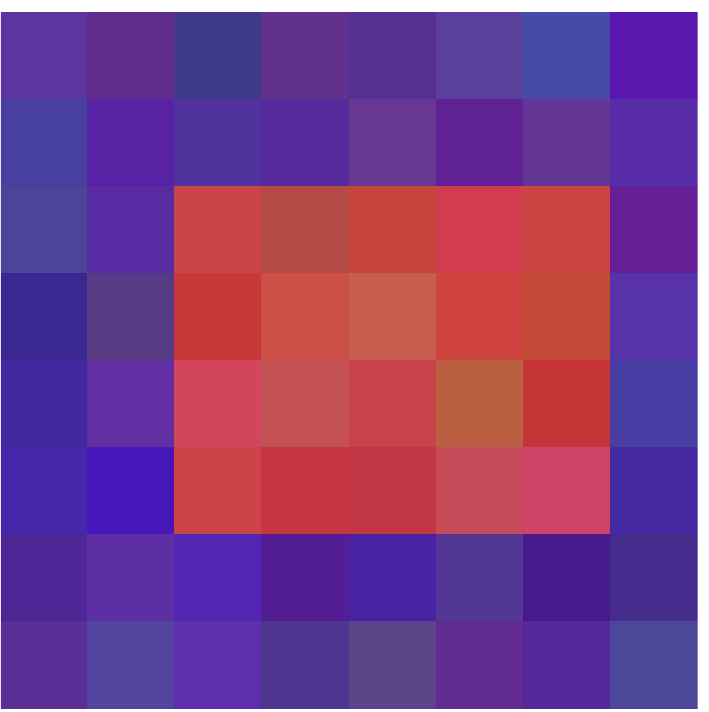}} \hspace*{0mm}
			\fbox{\includegraphics[width=0.45\linewidth]{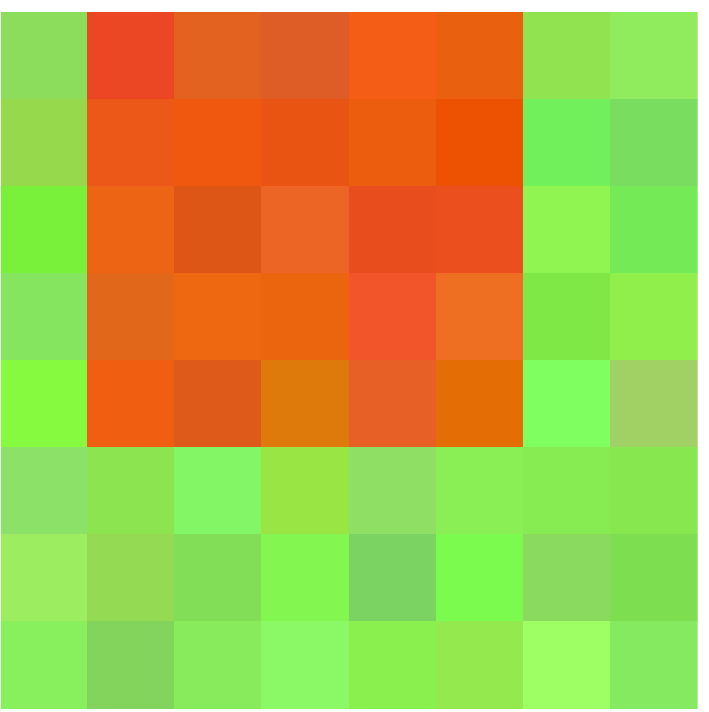}} \vspace*{-2.4mm} \\
			\fbox{\includegraphics[width=0.45\linewidth]{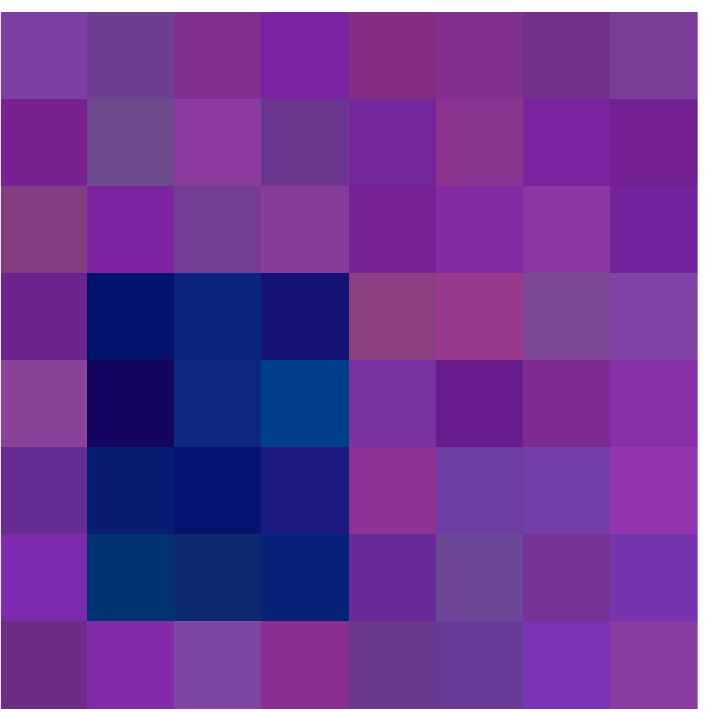}} \hspace*{0mm}
			\fbox{\includegraphics[width=0.45\linewidth]{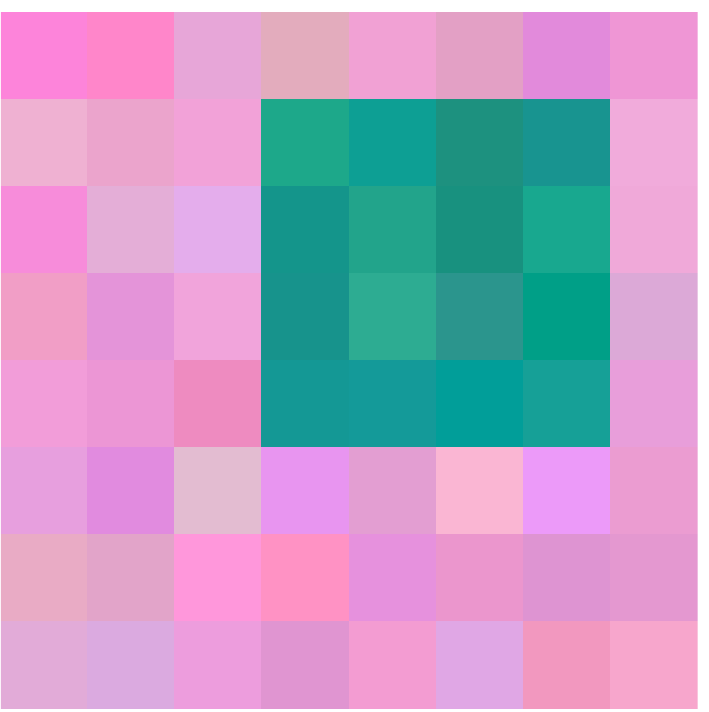}}
		}
		\hspace{0.2cm}
		\label{fig:square-data}
	}
	\hfill
	\subfigure[]{
		\begin{tikzpicture}

			\draw[red!20] (-2, 2.7) -- (2, 2.7);
			\draw[red, dotted] (-2, 2.7) -- (2, 2.7);

			\draw[red!20] (-2, 5.5) -- (2, 5.5);
			\draw[red, dotted] (-2, 5.5) -- (2, 5.5);

			\node[obs] 											(x)			{$\mathbf{x}_i$}; %
			\node[latent, above=of x]                			(z)			{$\mathbf{z}_i$}; %

			\factor[above=of x, yshift=1mm] 					{noise}		{left:Gaussian} {} {}; %
			\factor[above=of z, yshift=10mm] 					{gate}		{below left:Gate} {} {}; %

			\node[latent, right=of gate]                		(fg)		{$\mathrm{\mathbf{fg}}$}; %
			\node[latent, left=of gate]                			(bg)		{$\mathrm{\mathbf{bg}}$}; %
			\node[latent, above=of gate, yshift=-5mm]			(s)			{$s_i$}; %

			\factor[above=of s, yshift=10mm] 					{insq}		{below left, yshift=-2mm:Square} {} {}; %
			\factor[above=of fg] 								{pfg}		{} {} {}; %
			\factor[above=of bg] 								{pbg}		{} {} {}; %

			\node[latent, left=of insq]                			(c)			{$\mathbf{c}$}; %
			\node[latent, right=of insq]                		(l)			{$l$}; %
			\node[latent, above=of insq, yshift=-5mm, draw=none](p)			{$p_i$}; %

			\factor[above=of c] 								{pc}		{} {} {}; %
			\factor[above=of l] 								{pr}		{} {} {}; %

			\factoredge {z} 		{noise} 		{x}; %
			\factoredge {s} 		{gate} 			{z}; %
			\factoredge {c} 		{insq} 			{}; %
			\factoredge {l} 		{insq} 			{}; %
			\factoredge {p} 		{insq} 			{s}; %
			\factoredge {fg} 		{gate} 			{}; %
			\factoredge {bg} 		{gate} 			{}; %
			\factoredge {} 			{pfg} 			{fg}; %
			\factoredge {} 			{pbg} 			{bg}; %
			\factoredge {} 			{pc} 			{c}; %
			\factoredge {} 			{pr} 			{l}; %

			\plate {} {(p) (x)} {}; %

			\fill (bg.south) ++ (0, -0.52) circle (3pt) [fill=red] { };
			\fill (fg.south) ++ (0, -0.52) circle (3pt) [fill=red] { };

			\fill (l.south) ++ (0, -0.52) circle (3pt) [fill=red] { };

			\draw [-stealth, red] (-1.45, 2.85) -- (-1.45, 3.15);
			\draw [-stealth, red] (1.45, 2.85) -- (1.45, 3.15);

			\draw [-stealth, red] (1.45, 5.65) -- (1.45, 5.95);

			\node[yshift=-0.9cm] at (fg.south) { $\textcolor{red}{\Delta^{\mathrm{\mathbf{fg}}}}$ };
			\node[yshift=-0.9cm] at (bg.south) { $\textcolor{red}{\Delta^{\mathrm{\mathbf{bg}}}}$ };

			\node[yshift=-0.9cm] at (l.south) { $\textcolor{red}{\Delta^l}$ };)

		\end{tikzpicture}
		\hspace{0.1cm}
		\label{fig:square-model}
	}
	\mycaption{The square problem}{(a)~We wish to infer the square's center and its side length. (b)~A graphical model for this problem. $s_i$ is a boolean variable indicating the square's presence at position $p_i$. Depending on the value of $s_i$, the gate copies the appropriate colour ($\mathrm{\mathbf{fg}}$ or $\mathrm{\mathbf{bg}}$) to $\mathbf{z}_i$.}
	\label{fig:square}
\end{figure}

\begin{figure}[t]
	\centering
	\subfigure[Center]{
		\includegraphics[width=0.46\linewidth]{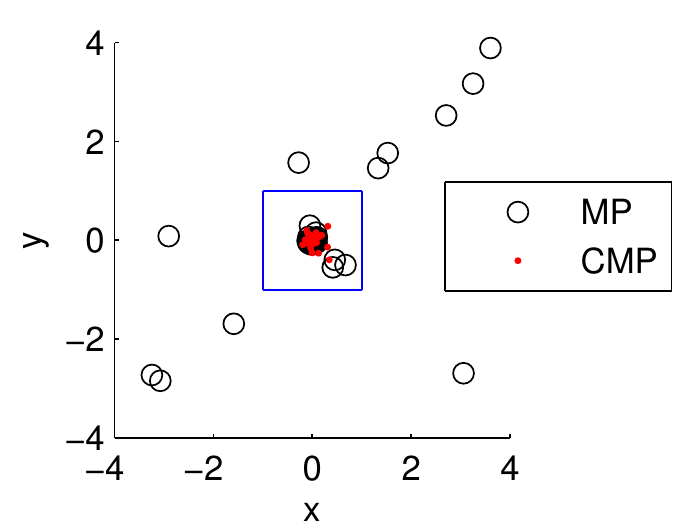}
		\label{fig:square-results-bullseye}
	}
	\subfigure[Center]{
		\includegraphics[width=0.46\linewidth]{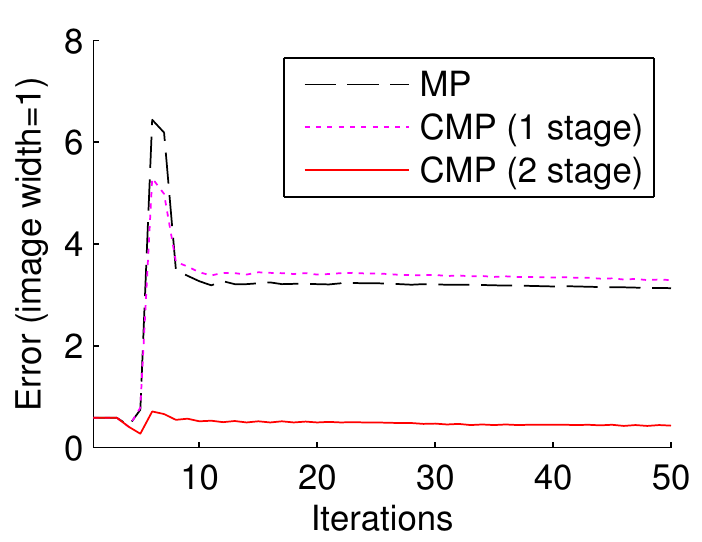}
		\label{fig:square-results-center}
	}
	\subfigure[Side length]{
		\includegraphics[width=0.46\linewidth]{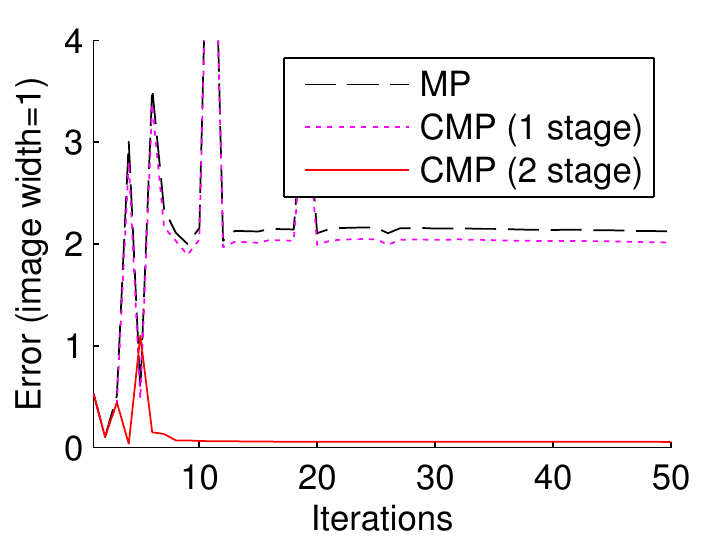}
		\label{fig:square-results-radius}
	}
	\subfigure[BG color]{
		\includegraphics[width=0.46\linewidth]{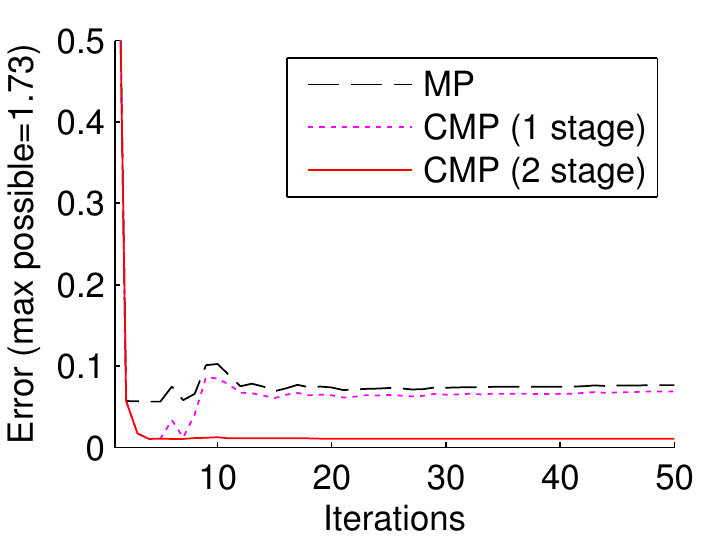}
		\label{fig:square-results-bgcolor}
	}
	\mycaption{Robustified inference using \MTD}{(a)~Position of inferred centers relative to groundtruth. Image boundaries shown in blue for scale. (b,c,d)~Distance of the mean of the posterior of $\mathbf{c}$, $l$ and $\mathrm{\mathbf{bg}}$ from their true values. \MTD consistently increases inference accuracy. Results have been averaged over 50 different problems. 1 stage \MTD only makes use of the lower predictors $\Delta^\mathrm{\mathbf{fg}}$ and $\Delta^\mathrm{\mathbf{bg}}$.}
	\label{fig:square-results}
\end{figure}

\subsection{A generative model of squares}
\label{sec:square}

Next we turn our attention to a more challenging problem for which even the best message passing scheme that we could devise frequently finds completely inaccurate solutions. The task is to infer the center $\mathbf{c}$ and side length $r$ of a square in an image (Fig.~\ref{fig:square-data}). Unlike the previous problem where we knew that all points belonged to the circle, here we must first determine which pixels belong to the square and which do not. To do so we might also wish to reason about the colour of the foreground $\mathrm{\mathbf{fg}}$ and background $\mathrm{\mathbf{bg}}$, making the task of inference significantly harder. The graphical model for this problem is shown in Fig.~\ref{fig:square-model}.

We experiment with 50 test images (themselves samples from the model), perform inference using EP and with a sequential schedule, recording the accuracy of the marginals of the latent variables at each iteration. We additionally place damping with step size 0.95 on messages from the square factor to the center $\mathbf{c}$. We found these choices led to the best performing standard message passing algorithm. Despite this, we observed inference accuracy to be disappointingly poor (see Fig.~\ref{fig:square-results}). In Fig.~\ref{fig:square-results-bullseye} we see that, for many images, message passing converges to highly inaccurate marginals for the center. The low quality of inference can also be seen in quantitative results of Figs.~\ref{fig:square-results}(b-d).

We implement \MTD predictors at two different layers of the model (see Fig.~\ref{fig:square-model}, red). In the first layer, $\Delta^\mathrm{\mathbf{fg}}$ and $\Delta^\mathrm{\mathbf{bg}}$ send consensus messages to $\mathrm{\mathbf{fg}}$ and $\mathrm{\mathbf{bg}}$ respectively, given the messages coming up from all of the $\mathbf{z}_i$ which take the form of independent Gaussians centered at the appearances of the observed pixels (we use a Gaussian noise model). Therefore $\Delta^\mathrm{\mathbf{fg}}$ and $\Delta^\mathrm{\mathbf{bg}}$ effectively make initial guesses of the values of the foreground and background colours in the image given the observed image. Split features in the internal nodes of the regression forest are designed to test for equality of two randomly chosen pixel positions, and sparse regressors are used at the leaves to prevent overfitting.

In the second layer, $\Delta^l$ sends a consensus message to $l$ given the messages coming up from all of the $s_i$. The messages from $s_i$ take the form of independent Bernoullis indicating the algorithm's current beliefs about the presence of the square at each pixel. Therefore the predictor's job is to predict the square's side length from this probabilistic segmentation map. Note that it is much easier to implement a regressor to perform this task (effectively one only needs to count) than it is to do so using the original observed image pixels $x_i$. We find these predictors to be sufficient for stable inference and so we do not implement a fourth predictor for $\mathbf{c}$. We experiment with single stage \MTD, where only the lower predictors $\Delta^\mathrm{\mathbf{fg}}$ and $\Delta^\mathrm{\mathbf{bg}}$ are active, and with two stage \MTD, where all three predictors are active. The predictors are trained on $D=500$ samples from the model.

The results of these experiments are shown in Fig.~\ref{fig:square-results}. We observe that \MTD significantly improves the accuracy of inference for the center $\mathbf{c}$ (Figs.~\ref{fig:square-results-bullseye}, \ref{fig:square-results-center}) but also for the other latent variables (Figs.~\ref{fig:square-results-radius}, \ref{fig:square-results-bgcolor}). Of note is the fact that single stage \MTD appears to be insufficient for guiding message passing to good solutions. Whereas in circle example \MTD accelerated convergence, this example demonstrates how it can make inference possible in models that were outside the capabilities of standard message passing.

\subsection{A generative model of faces}
\label{sec:shading}

\begin{figure}[t]
	\centering
	\subfigure[]{
		\setlength\fboxsep{-0.3mm}
		\setlength\fboxrule{0pt}
		\parbox[b]{3cm}{
			\centering
			\small
			\fbox{\includegraphics[width=1.3cm]{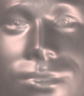}}\\
			Normal $\{\mathbf{n}_i \}$ \vspace{4.5mm} \\
			\fbox{\includegraphics[width=1.3cm]{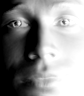}} \\
			Shading $\{ s_i \}$ \vspace{4.5mm} \\
			\fbox{\includegraphics[width=1.3cm]{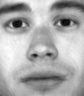}} \\
			Reflectance $\{ r_i \}$ \vspace{4.5mm} \\
			\fbox{\includegraphics[width=1.3cm]{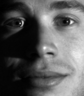}} \\
			Observed image $\{ x_i \}$\\
		}
		\hspace{0.1cm}
		\label{fig:shading-data}
	}
	\hfill
	\subfigure[]{
		\begin{tikzpicture}

			\draw[red!20] (-2.4, 2.7) -- (2, 2.7);
			\draw[red, dotted] (-2.4, 2.7) -- (2, 2.7);

			\draw[red!20] (-2.4, 5.5) -- (2, 5.5);
			\draw[red, dotted] (-2.4, 5.5) -- (2, 5.5);

			\node[obs] 											(x)			{$x_i$}; %
			\node[latent, above=of x]                			(z)			{$z_i$}; %

			\factor[above=of z, yshift=10mm] 					{times}		{below left: $\times$} {} {}; %

			\node[latent, above=of times, yshift=-5mm]			(s)			{$s_i$}; %

			\factor[above=of x, yshift=1mm] 					{noise1}	{left:Gaussian} {} {}; %

			\node[latent, left=of times]                		(r)			{$r_i$}; %

			\factor[above=of r] 								{pr}		{} {} {}; %

			\factor[above=of s, yshift=10mm] 					{inner}		{left:Product} {} {}; %

			\node[latent, above=of inner, yshift=-5mm] 			(n)			{$\mathbf{n_i}$}; %
			\node[latent, right=of inner]                		(l)			{$\mathbf{l}$}; %

			\factor[above=of l] 								{pl}		{} {} {}; %

			\factor[above=of n] 								{pn}		{} {} {}; %

			\factoredge {z} 		{noise1} 		{x}; %
			\factoredge {s} 		{times} 		{z}; %
			\factoredge {r} 		{times} 		{}; %
			\factoredge {n} 		{inner} 		{s}; %
			\factoredge {l} 		{inner} 		{}; %
			\factoredge {} 			{pn} 			{n}; %
			\factoredge {} 			{pl} 			{l}; %
			\factoredge {} 			{pr} 			{r}; %

			\plate {} {(pn) (x) (r)} {}; %

			\fill (r.south) ++ (0, -0.52) circle (3pt) [fill=red] { };

			\fill (l.south) ++ (0, -0.52) circle (3pt) [fill=red] { };

			\draw [-stealth, red] (-1.45, 2.85) -- (-1.45, 3.15);

			\draw [-stealth, red] (1.45, 5.65) -- (1.45, 5.95);

			\node[yshift=-0.9cm] at (r.south) { $\textcolor{red}{\Delta_i^\mathbf{r}}$ };

			\node[yshift=-0.9cm] at (l.south) { $\textcolor{red}{\Delta^\mathbf{l}}$ };

			\node[yshift=0.35cm, xshift=-0.83cm] at (inner) { Inner };			

		\end{tikzpicture}
		\label{fig:shading-model}
	}
	\mycaption{The face problem}{(a)~We observe an image and wish to infer the corresponding reflectance map and normal map (visualized here as 3D shape). (b)~A graphical model for this problem. Symmetry priors not shown.}
	\label{fig:shading}
	\vspace*{-4mm}
\end{figure}

We also investigate a more realistic application to face modelling. The estimation of reflectance and shape from a single image of a human face is a well-studied problem in computer vision (see \eg \citealt{Georghiades2001, Lee2005, Wang2009, Kemelmacher2011, Tang2012}). A primary motivation for this task is that reflectance and shape are invariant to confounding light effects, and are therefore useful for downstream tasks such as recognition. The problem is ill-posed however, and modern approaches make heavy use of prior knowledge in order to obtain good solutions, \eg in the form of average reflectance and normal statistics \citep{Biswas2009, Biswas2010} or morphable 3D models \citep{Zhang2006, Wang2009}.

\textbf{Model.} Given an observation of pixels $\mathbf{x} = \{x_i\}$, we wish to infer the reflectance value $r_i$ and normal vector $\mathbf{n_i}$ for each pixel $i$ (see Fig.~\ref{fig:shading-data}). In Fig.~\ref{fig:shading-model}, a model is shown for these variables that represents the following image formation process: $x_i = (\mathbf{n_i} \cdot \mathbf{l}) \times r_i + \epsilon$, thereby assuming Lambertian reflection and an infinitely distant directional light source with variable intensity. We place Gaussian priors over reflectances $\{ r_i \}$, normals $\{ \mathbf{n_i} \}$, and the light $\mathbf{l}$; and set the parameters of the priors using training data. We additionally place a soft symmetry prior on the $\{ r_i \}$ (the reflectance value on one side of the face should be close to its value on the other side) and on the $\{ \mathbf{n_i} \}$ (normal vectors on each side should be approximately symmetric), reflecting our prior knowledge about faces. These symmetry priors can be added to the model in just a few lines of code, illustrating the way in which model-based methods lend themselves to rapid prototyping and experimentation.

Although this model is only a crude approximation to the true image formation process (\eg it does not account for shadows or specularities), similar approximations have been found to be useful in prior work \citep{Biswas2009, Biswas2010, Kemelmacher2011}. Additionally, if we can successfully develop algorithms that perform accurate and reliable inference in this class of models, we would then be able to increase its usefulness simply by updating it to reflect the true image formation process more accurately. Note that even for a relatively small image of size $96 \times 84$, the model contains over 48,000 latent variables and 56,000 factors, and as we will show below, standard message passing in the model routinely fails to converge to accurate solutions.

\begin{figure*}
	\centering
	\setlength\fboxsep{0.2mm}
	\setlength\fboxrule{0pt}
	\begin{tikzpicture}
	
	\node at (-7.3, 2.9) {\small (a) Observed};
	\node at (-4.2, 2.9) {\small (b) Reflectance};
	\node at (-0.58, 2.9) {\small (c) Variance};
	\node at (2.35, 2.9) {\small (d) Light};
	\node at (6.4, 2.9) {\small (e) Normal};
	
	\node at (-4.15, 2.6) {\tiny $\color{gray}{\overbrace{\quad\quad\quad\quad\quad\quad\quad\quad\quad\quad
														  \quad\quad\quad\quad\quad\quad\quad\quad\quad\quad
														  \quad\quad\quad\quad\quad\quad}}$};
															  
    \node at (2.45, 2.6) {\tiny $\color{gray}{\overbrace{\quad\quad\quad\quad\quad\quad\quad\quad\quad\quad
														 \quad\quad\quad\quad\quad\quad\quad\quad\quad\quad}}$};
															  
	\node at (6.58, 2.6) {\tiny $\color{gray}{\overbrace{\quad\quad\quad\quad\quad\quad\quad\quad\quad\quad
														 \quad\quad\quad\quad\quad\,\,\,}}$};

		\matrix at (0, 0) [matrix of nodes, nodes={anchor=east}, column sep=-0.15cm, row sep=-0.2cm]
		{
			\fbox{\includegraphics[width=1cm]{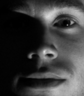}} & \hspace{0.12cm} & 
			\fbox{\includegraphics[width=1cm]{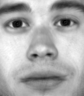}} &
			\fbox{\includegraphics[width=1cm]{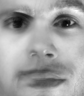}} &
			\fbox{\includegraphics[width=1cm]{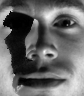}} &
			\fbox{\includegraphics[width=1cm]{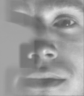}}  &
			\fbox{\includegraphics[width=1cm]{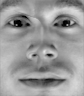}} & \hspace{0.12cm} & 
			\fbox{\includegraphics[width=1cm]{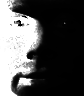}} & \hspace{0.12cm} & 
			\fbox{\includegraphics[width=1cm]{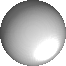}} &
			\fbox{\includegraphics[width=1cm]{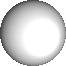}} &
			\fbox{\includegraphics[width=1cm]{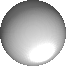}}  &
			\fbox{\includegraphics[width=1cm]{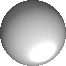}} & \hspace{0.12cm} & 
			\fbox{\includegraphics[width=1cm]{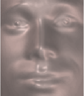}} &
			\fbox{\includegraphics[width=1cm]{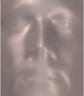}} &
			\fbox{\includegraphics[width=1cm]{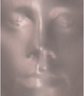}}  	\\
			
			\fbox{\includegraphics[width=1cm]{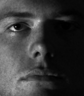}} & \hspace{0.12cm} & 
			\fbox{\includegraphics[width=1cm]{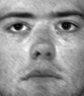}} &
			\fbox{\includegraphics[width=1cm]{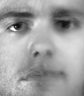}} &
			\fbox{\includegraphics[width=1cm]{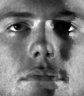}} &
			\fbox{\includegraphics[width=1cm]{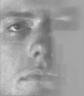}} &
			\fbox{\includegraphics[width=1cm]{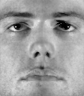}} & \hspace{0.12cm} & 
			\fbox{\includegraphics[width=1cm]{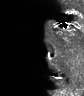}} & \hspace{0.12cm} & 
			\fbox{\includegraphics[width=1cm]{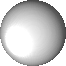}} &
			\fbox{\includegraphics[width=1cm]{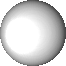}}  &
			\fbox{\includegraphics[width=1cm]{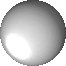}} &
			\fbox{\includegraphics[width=1cm]{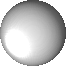}} & \hspace{0.12cm} & 
			\fbox{\includegraphics[width=1cm]{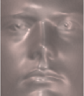}} &
			\fbox{\includegraphics[width=1cm]{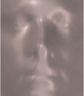}} &
			\fbox{\includegraphics[width=1cm]{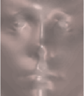}}  \\

			\fbox{\includegraphics[width=1cm]{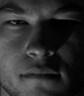}} & \hspace{0.12cm} & 
			\fbox{\includegraphics[width=1cm]{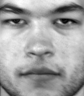}} &
			\fbox{\includegraphics[width=1cm]{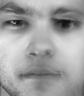}} &
			\fbox{\includegraphics[width=1cm]{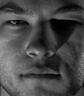}} &
			\fbox{\includegraphics[width=1cm]{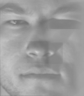}} &
			\fbox{\includegraphics[width=1cm]{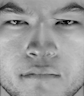}} & \hspace{0.12cm} & 
			\fbox{\includegraphics[width=1cm]{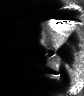}} & \hspace{0.12cm} & 
			\fbox{\includegraphics[width=1cm]{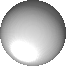}} &
			\fbox{\includegraphics[width=1cm]{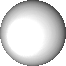}} &
			\fbox{\includegraphics[width=1cm]{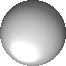}}  &
			\fbox{\includegraphics[width=1cm]{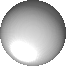}} & \hspace{0.12cm} & 
			\fbox{\includegraphics[width=1cm]{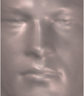}} &
			\fbox{\includegraphics[width=1cm]{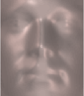}} &
			\fbox{\includegraphics[width=1cm]{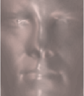}} \\
			
			\fbox{\includegraphics[width=1cm]{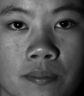}} & \hspace{0.12cm} & 
			\fbox{\includegraphics[width=1cm]{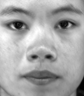}} &
			\fbox{\includegraphics[width=1cm]{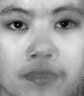}} &
			\fbox{\includegraphics[width=1cm]{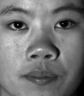}} &
			\fbox{\includegraphics[width=1cm]{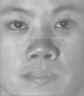}} &
			\fbox{\includegraphics[width=1cm]{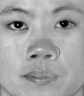}} & \hspace{0.12cm} & 
			\fbox{\includegraphics[width=1cm]{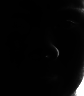}} & \hspace{0.12cm} & 
			\fbox{\includegraphics[width=1cm]{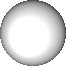}} &
			\fbox{\includegraphics[width=1cm]{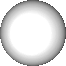}} &
			\fbox{\includegraphics[width=1cm]{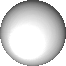}} &
			\fbox{\includegraphics[width=1cm]{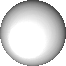}} & \hspace{0.12cm} & 
			\fbox{\includegraphics[width=1cm]{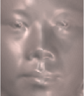}} &
			\fbox{\includegraphics[width=1cm]{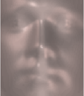}} &
			\fbox{\includegraphics[width=1cm]{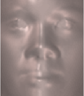}}  \\
	     };
	     
	     \node at (-6.4, -2.7) {\small `GT'};
	     \node at (-5.25, -2.7) {\small BU};
	     \node at (-4.1, -2.7) {\small MP};
	     \node at (-3.0, -2.7) {\small Forest};
	     \node at (-1.87, -2.7) {\small \textbf{CMP}};

	     \node at (-0.55, -2.7) {\small \textbf{CMP}};
	     
	     \node at (0.75, -2.7) {\small `GT'};
	     \node at (1.9, -2.7) {\small MP};
	     \node at (3, -2.7) {\small Forest};
	     \node at (4.15, -2.7) {\small \textbf{CMP}};
	     
	     \node at (5.5, -2.7) {\small `GT'};
	     \node at (6.6, -2.7) {\small MP};
	     \node at (7.72, -2.7) {\small \textbf{CMP}};

	\end{tikzpicture}
	\mycaption{A visual comparison of inference results}{For 4 randomly chosen test images, we show inference results obtained by competing methods. (a)~Observed images. (b)~Inferred reflectance maps. \textit{GT} is the stereo estimate which we use as a proxy for groundtruth, \textit{BU} is the bottom-up reflectance estimate of Biswas \etal (2009) and \textit{Forest} is the consensus prediction. (c)~The variance of the inferred reflectance estimate produced by \MTD (normalized across rows). High variance regions correlate strongly with cast shadows. (d)~Visualization of inferred light. (e)~Inferred normal maps.}
	\label{fig:shading-qualitative-multiple-subjects}
\end{figure*}

\textbf{\Method.} We use predictors at two levels in the model (see Fig.~\ref{fig:shading-model}) to tackle this problem. The first sends consensus messages to \textit{each} reflectance pixel $r_i$, making it an instance of type B of \MTD as described in Fig.~\ref{fig:types-b}. Here, each consensus message is predicted using information from all the contextual messages from the $z_i$. We denote each of these predictors by $\Delta_i^\mathbf{r}$. The second predictor sends a consensus message to $\mathbf{l}$ using information from all the messages from the $s_i$ and is denoted by $\Delta^\mathbf{l}$. The first level of predictors effectively make a guess of the reflectance image from the denoised observation, and the second layer predictor produces an estimate of the light from the shading image (which is likely to be easier to do than directly from the observation). The reflectance predictors $\{ \Delta_i^\mathbf{r} \}$ are all powered by a single random forest, however the pixel position $i$ is used as a feature that it can exploit to create location specific behaviour. The tree parameterization of the contextual messages $\mathbf{c}$ for use in the reflectance predictor $\Delta_i^\mathbf{r}$ also includes 16 features such as mean, median, max, min and gradients of a $21 \times 21$ patch around the pixel. The tree parameterization of the contextual messages for use in the lighting predictor $\Delta^\mathbf{l}$ consists of means of the mean of the shading messages in $12 \times 12$ blocks. We deliberately use simple features to maintain generality but one could imagine the use of more specialized regressors for maximal performance.

\textbf{Datasets.} We experiment with the `Yale B' and `Extended Yale B' datasets~\citep{Georghiades2001, Lee2005}. Together, they contain images of 38 subjects each with 64 illumination directions. We remove images taken with extreme light angles (azimuth or elevation $\ge 85$ degrees) that are almost entirely in shadow, leaving around 45 images for each subject. Images are downsampled to $96 \times 84$. There are no groundtruth normals or reflectances for this dataset, however it is common practice to create proxy groundtruths using photometric stereo, which we do using the code of~\cite{Queau2013}. We use images from 22 subjects for training and test on the remaining 16 subjects.

\textbf{Results.} We begin by qualitatively assessing the different inference schemes. In Fig.~\ref{fig:shading-qualitative-multiple-subjects} we show inference results for reflectance maps, normal maps and lights that are obtained following 100 iterations of message passing (VMP). For reflectance (Fig.~\ref{fig:shading-qualitative-multiple-subjects}b), we would like inference to produce estimates that match closely the groundtruth produced by photometric stereo (GT). We also display the reflectance estimates produced by the strong baseline of \cite{Biswas2009} for reference. We note that the baseline achieves excellent accuracy in regions with strong lighting, however it produces blurry estimates in regions under shadow.

As can be seen in Fig.~\ref{fig:shading-qualitative-multiple-subjects}b (MP), standard variational message passing finds solutions that are highly inaccurate with continued presence of illumination and artefacts in areas of cast show. In contrast, inference using \MTD produces artefact-free results that much more closely resemble the stereo groundtruths. Arguably \MTD also improves over the baseline \citep{Biswas2009}, since its estimates are not blurry in regions with cast shadows. This is made possible by the presence of symmetry priors in the model. Additionally, we note that the variance of the \MTD inference for reflectance (Fig.~\ref{fig:shading-qualitative-multiple-subjects}c) correlates strongly with cast shadows in the observed images (\ie the model is uncertain where it should be) suggesting that in future work it would be fruitful to have the notion of cast shadows explicitly built into the model. Figs.~\ref{fig:shading-qualitative-multiple-subjects}d and \ref{fig:shading-qualitative-multiple-subjects}e show analogous results for lighting and normal maps, and Fig.~\ref{fig:shading-qualitative-same-subject} demonstrates \MTD's ability to robustly infer reflectance maps for images of a single subject taken under varying lighting conditions.

We use the task of subject recognition (using estimated reflectance) as a quantitative measure of inference accuracy, as it can be difficult to measure in more direct ways (\eg RMSE strongly favours blurry predictions). The reflectance estimate produced by each algorithm is compared to all training subjects' groundtruth reflectances and is assigned the label of its closest match. We have found this evaluation to reflect the quality of inference and we choose to use it for its simplicity. Fig.~\ref{fig:shading-quantitative-reflectance} shows the result of this experiment, both for real images and also synthetic images that were produced by taking the stereo groundtruths and adding artificial lighting (but with no cast shadows). We show analogous results for light in Fig.~\ref{fig:shading-quantitative-light}, where error is defined to be the cosine angle distance between the estimated light and the photometric stereo reference. First, we note that standard variational message passing (MP) performs poorly, producing reflectance estimates that are much less useful for recognition than those from \cite{Biswas2009}. Second, we note that \MTD in the same model (both 1 stage and 2 stage versions) produces inferences that are significantly more useful downstream. The horizontal line labelled `Forest' represents the accuracy of the consensus messages without any message passing, showing that the model-based fine-tuning provides a significant benefit. Finally, we highlight the fact that initializing light directly from the image and running message passing (Fig.~\ref{fig:shading-quantitative-reflectance}, Init+MP) leads to worse estimates than \MTD demonstrating the use of layered predictions as opposed to direct predictions from the observations. These results demonstrate that \MTD helps message passing find better fixed points even in the presence of model mis-match (shadows) and make use of the full potential of the generative model.

\begin{figure}
	\centering
	\setlength\fboxsep{0.2mm}
	\setlength\fboxrule{0pt}
	\begin{tikzpicture}

		\matrix at (0, 0) [matrix of nodes, nodes={anchor=east}, column sep=-0.15cm, row sep=-0.2cm]
		{
			\fbox{\includegraphics[width=1cm]{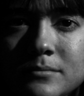}} &
			\fbox{\includegraphics[width=1cm]{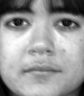}} &
			\fbox{\includegraphics[width=1cm]{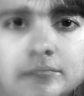}}  &
			\fbox{\includegraphics[width=1cm]{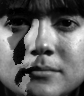}}  &
			\fbox{\includegraphics[width=1cm]{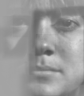}}  &
			\fbox{\includegraphics[width=1cm]{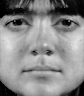}}  &
			\fbox{\includegraphics[width=1cm]{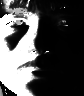}} 
			 \\
			
			\fbox{\includegraphics[width=1cm]{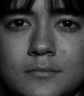}} &
			\fbox{\includegraphics[width=1cm]{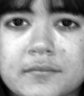}} &
			\fbox{\includegraphics[width=1cm]{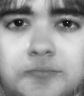}}  &
			\fbox{\includegraphics[width=1cm]{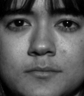}}  &
			\fbox{\includegraphics[width=1cm]{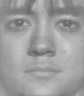}}  &
			\fbox{\includegraphics[width=1cm]{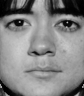}}  &
			\fbox{\includegraphics[width=1cm]{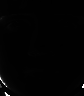}} 
			 \\

			\fbox{\includegraphics[width=1cm]{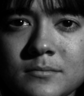}} &
			\fbox{\includegraphics[width=1cm]{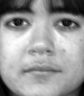}} &
			\fbox{\includegraphics[width=1cm]{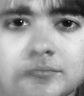}} &
			\fbox{\includegraphics[width=1cm]{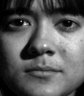}}  &
			\fbox{\includegraphics[width=1cm]{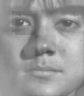}}  &
			\fbox{\includegraphics[width=1cm]{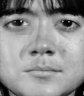}}  &
			\fbox{\includegraphics[width=1cm]{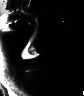}} 
			 \\
	     };

	     \node at (-3.38, -2.0) {\small Observed};
	     \node at (-2.25, -2.0) {\small `GT'};
	     \node at (-1.15, -2.0) {\small BU};
	     \node at (0, -2.0) {\small MP};
	     \node at (1.15, -2.0) {\small Forest};
	     \node at (2.25, -2.0) {\small \textbf{CMP}};
	     \node at (3.38, -2.0) {\small Variance};

	\end{tikzpicture}
	\mycaption{Robustness to varying illumination}{Left to right: observed image, photometric stereo estimate (proxy for groundtruth), \cite{Biswas2009} estimate, VMP result, consensus forest estimate, CMP mean, and CMP variance.}
	\label{fig:shading-qualitative-same-subject}
\end{figure}

\begin{figure}[t]
	\centering
	\subfigure[Without shadows]{
		\includegraphics[width=0.46\linewidth]{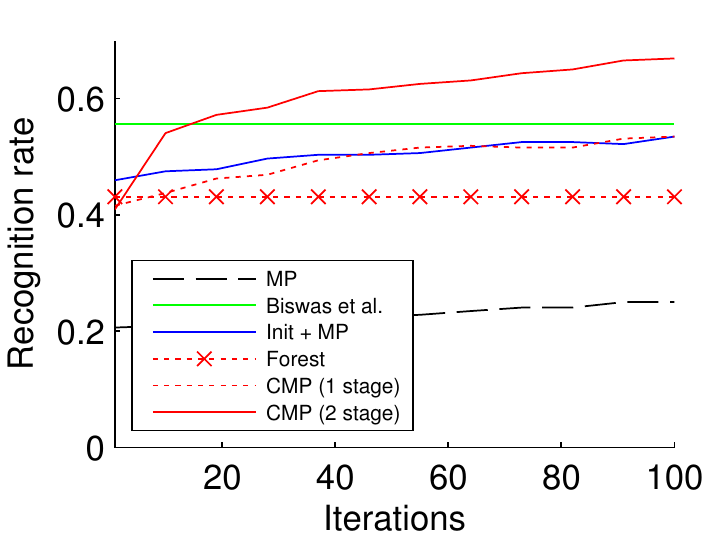}
	}
	\subfigure[With shadows]{
		\includegraphics[width=0.46\linewidth]{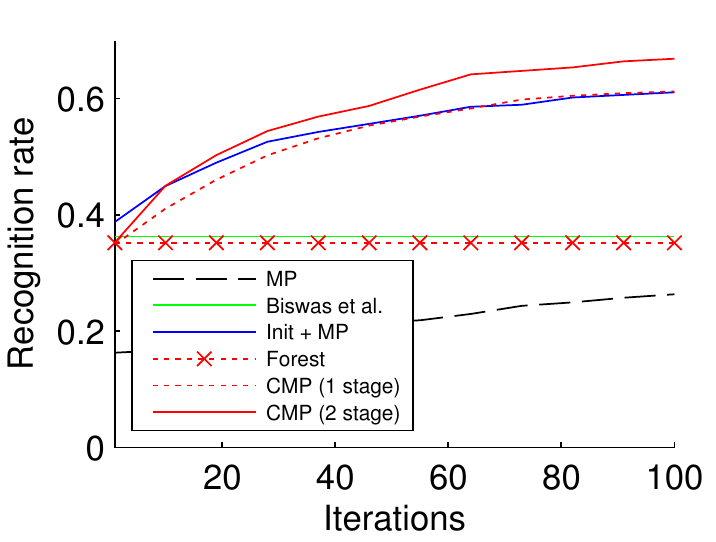}
	}
	\mycaption{Reflectance inference accuracy demonstrated through recognition accuracy}{\MTD allows us to make use of the full potential of the generative model, thereby outperforming the competitive bottom-up method of \cite{Biswas2009}.}
	\label{fig:shading-quantitative-reflectance} 

	\subfigure[Without shadows]{
		\includegraphics[width=0.46\linewidth]{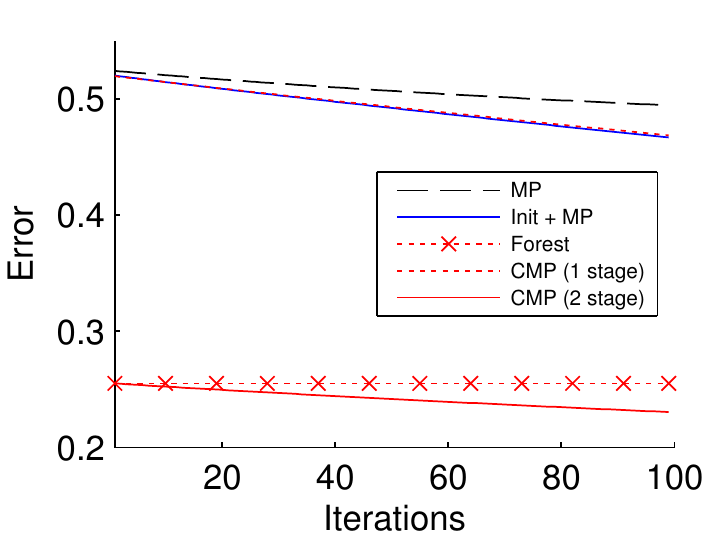}
	}
	\subfigure[With shadows]{
		\includegraphics[width=0.46\linewidth]{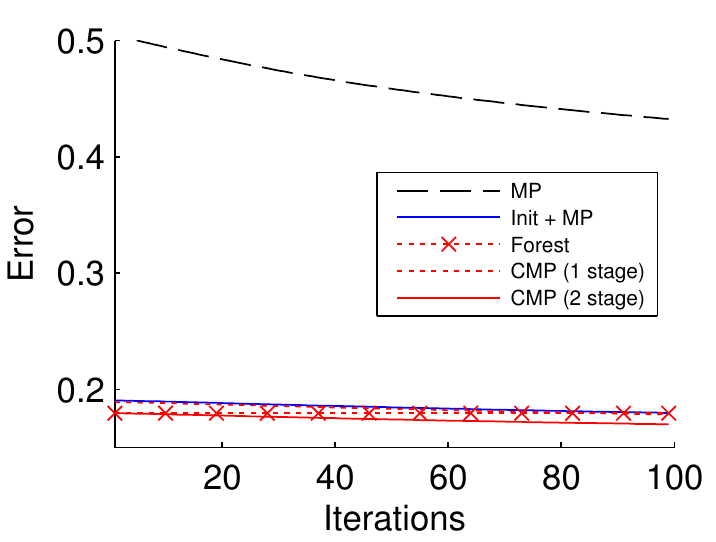}
	}
	\mycaption{Light inference accuracy}{The presence of cast shadows makes the direct prediction task easier, however \MTD is accurate even in their absence.}
	\label{fig:shading-quantitative-light}
\end{figure}

\vspace{-0.1cm}
\section{Related Work}
\label{sec:related-work}
\vspace{-0.2cm}

Inspiration for \MTD stems from the kinds of distinctions that have been made for decades between so-called `intuitive', bottom-up, fast inference techniques, and iterative `rational' inference techniques~\citep{Hinton1990}. \MTD can be seen as an implementation of such ideas in the context of message passing, where the consensus messages form the `intuitive' part of inference and the following standard message passing forms the `rational' part. Analogues to intuitive and rational inference also exist for sampling, where bottom-up techniques are used to compute proposals for MCMC, leading to significant speedup in inference~\citep{Tu2001, Stuhlmuller2013, Jampani2014}. \cite{rezende2014stochastic} and \cite{kingma2013auto} proposed techniques for learning the parameters of both the generative model and the corresponding recognition model.

The idea of `learning to infer' also has a long history. Early examples include~\cite{Hinton1995}, where a dedicated set of `recognition' parameters are learned to drive inference. In more modern instances of such ideas \citep{Munoz2010, Ross2011, Domke2011, Shapovalov2013, Munoz2013}, message passing is performed  by a sequence of predictions defined by a graphical model, and the predictors are jointly trained to ensure that the system produces correct labellings. However in these techniques the resulting inference procedure no longer corresponds to the original (or perhaps to any) graphical model. An important distinction of \MTD is that the predictors fit completely within the framework of message passing and final inference results correspond to valid fixed points in the original model of interest.

Finally, we note recent works of~\cite{Heess2013} and~\cite{Eslami2014} that make use of regressors (neural networks and random forests, respectively) to learn to pass EP messages. These works are concerned with reducing the computational cost of computing individual messages and do not make any attempt to change the accuracy or rate of convergence in message passing inference as a whole. In contrast, \MTD learns to pass messages specifically with the aim of reducing the total number of iterations required for accurate inference in a given generative model.

\vspace{-0.1cm}
\section{Discussion}
\label{sec:discussion}
\vspace{-0.2cm}

We have presented \METHOD and shown that it is a computationally efficient technique that can be used to improve the accuracy of message passing inference in a variety of vision models. The crux of the approach is to recognize the importance of global variables, and to take advantage of layered model structures commonly seen in vision to make rough estimates of their values.

The success of \MTD depends on the accuracy of the random forest predictors. The design of forest features is not yet completely automated, but we took care in this work to use generic features that can be applied to a broad class of problems. Our forests are implemented in an extensible manner, and we envisage building a library of them that one can choose from, simply by inspecting the data types of the contextual and target variables. 

In future work, we would like to exploit the benefits of the \MTD framework by applying it to more challenging problems from computer vision. Each of the examples in Sec.~\ref{sec:experiments} can be extended in various ways, \eg by making considerations for multiple objects, incorporating occlusion in the squares example and cast shadows in the faces example, or by developing more realistic priors. We are also seeking to understand in what other domains the application of our ideas may be fruitful.

More broadly, a major challenge in machine learning is that of enriching models in a scalable way. We continually seek to ask our models to provide interpretations of increasingly complicated, heterogeneous data sources. Graphical models provide an appealing framework to manage this complexity, but the difficulty of inference has long been a barrier to achieving these goals. The \MTD framework takes us one step in the direction of overcoming this barrier.

\small
\textbf{Acknowledgements.} We thank the anonymous reviewers, Tom Minka, Christopher Williams, Peter Gehler, Sebastian Nowozin and Andrew Fitzgibbon for their feedback and suggestions.
\normalsize

\bibliographystyle{apalike}
\bibliography{bibliography}


\onecolumn

%

%

%
%
%

\vspace{-15.0cm}

\aistatstitle{Consensus Message Passing for Layered Graphical Models \\ \textit{Supplementary Material}}

\aistatsauthor{ Varun Jampani\footnotemark[2] \And S. M. Ali Eslami\footnotemark[2], Daniel Tarlow, Pushmeet Kohli and John Winn }

\aistatsaddress{ MPI for Intelligent Systems, T\"{u}bingen \And Microsoft Research, Cambridge }

\footnotetext[2]{The first two authors contribute equally to this work.}

\setcounter{section}{0}

\section{Random regression forests for CMP}

We wish to learn a mapping $f$ from contextual messages $\mathbf{c}$ to the consensus message $m$ from training data $\{ (\mathbf{c}_d, m_d) \}_{d=1...D}$. This is challenging since the inputs and outputs of the regression problem are both messages (\ie distributions), and special care needs to be taken to account for this fact. We follow closely the methodology of~\cite{Eslami2014}, who use random forests to predict outgoing messages from a factor given the incoming messages to it. For a review of forests see~\citep{Criminisi2013}.

In approximate message passing (\eg EP;~\citealp{Minka2001} and VMP;~\citealp{Winn2005}), messages can be represented using only a few numbers, \eg a Gaussian message can be represented by its natural parameters. We represent the contextual messages $\mathbf{c}$ collectively, in two different ways: the first is a concatenation of the parameters of its constituent messages which we call the `regression parameterization' and denote by $\mathbf{r}_\textrm{c}$; and the second is a vector of features computed on the set which we call the `tree parameterization' and denote by $\mathbf{t}_\textrm{c}$. This parametrization typically contains features of the set as a whole (\eg moments of their means). We represent the outgoing message $m$ by a vector of real valued numbers $\mathbf{r}_\textrm{m}$.

\textbf{Prediction model.} Each leaf node is associated with a subset of the labelled training data. During testing, a previously unseen set of contextual messages represented by $\mathbf{t}_\textrm{c}$ traverses the tree until it reaches a leaf which by construction is likely to contain similar training examples. We therefore use the statistics of the data gathered in that leaf to predict the consensus message with a multivariate regression model of the form: $\mathbf{r}_\textrm{m} = \mathbf{W} \cdot \mathbf{r}_\textrm{c} + \epsilon$ where $\epsilon$ is a vector of normal error terms. We use the learned matrix of coefficients $\mathbf{W}$ at test time to make predictions $\overline{\mathbf{r}}_\textrm{m}$ for each $\mathbf{r}_\textrm{c}$. To recap, $\mathbf{t}_\textrm{c}$ is used to traverse the contextual messages down to leaves, and $\mathbf{r}_\textrm{c}$ is used by a linear regressor to predict the parameters $\mathbf{r}_\textrm{m}$ of the consensus message.

\textbf{Training objective function.} The optimization of the split functions proceeds in a greedy manner. At each node $j$, depending on the subset of the incoming training set $\mathcal{S}_j$ we learn the function that `best' splits $\mathcal{S}_j$ into the training sets corresponding to each child, $\mathcal{S}_j^\textrm{L}$ and $\mathcal{S}_j^\textrm{R}$, \ie the parameters of the split criterion $\boldsymbol{\tau}_j = \argmax_{\boldsymbol{\tau} \in \mathcal{T}_j} I(\mathcal{S}_j, \boldsymbol{\tau})$. This optimization is performed as a search over a discrete set $\mathcal{T}_j$ of a random sample of possible parameter settings. The objective function $I$ is:
\begin{align}
I(\mathcal{S}_j, \boldsymbol{\tau}) &= - E(\mathcal{S}_j^\textrm{L}, \mathbf{W}^\textrm{L}) - E(\mathcal{S}_j^\textrm{R}, \mathbf{W}^\textrm{R}),
\end{align}
where $\mathbf{W}^\textrm{L}$ and $\mathbf{W}^\textrm{R}$ are the parameters of the regression models corresponding to the left and right training sets $\mathcal{S}_j^\textrm{L}$ and $\mathcal{S}_j^\textrm{R}$, and $E$ is the `fit residual' as defined in~\citep{Eslami2014}. In simple terms, this objective function splits the training data at each node in a way that the relationship between the incoming and outgoing messages is well captured by the regression in each child.

\textbf{Ensemble model.} During testing, a set of contextual messages simultaneously traverses every tree in the forest from their roots until it reaches their leaves. Combining the predictions into a single forest prediction might be done by averaging the parameters $\overline{\mathbf{r}}_\textrm{m}^t$ of the predicted messages $\overline{m}^t$ by each tree $t$, however this would be sensitive to the chosen parameterization. Instead we compute the moment average $\overline{m}$ of the distributions $\{ \overline{m}^t \}$ by averaging the first few moments of the predictions across trees, and solving for the distribution parameters which match the averaged moments (see \eg \citealp{Grosse2013}). 

\clearpage

\section{Results on the face problem}

\subsection{Qualitative results}

Figure~\ref{fig:shading-qualitative-multiple-subjects} shows inference results for reflectance maps, normal maps and lights for randomly chosen test images, and Figure~\ref{fig:shading-qualitative-same-subject} shows reflectance estimation results on multiple images of the same subject produced under different illumination conditions. \Method is able to produce reflectance estimates that are closer to the photometric stereo groundtruth across subjects and across different illumination conditions.

\subsection{Quantitative results}

Figure~\ref{fig:shading-quantitative-reflectance} shows quantitative results for both real images from `Yale B' and `Extended Yale B' datasets \citep{Georghiades2001, Lee2005} and synthetic shadowless images. The synthetic shadowless images were created using the same light, reflectance and normal map statistics as that of images in the real dataset (however estimated using photometric stereo~\citep{Queau2013}). Subject recognition results indicate superior performance of \MTD in comparison to other baselines in both real and synthetic image settings.

Figure~\ref{fig:shading-quantitative-light} shows the quantitative results of light inference using the different inference techniques. We use the cosine angle distance between the estimated light and the photometric stereo groundtruth ($\textrm{error} = \cos^{-1}(\hat{\mathbf{l}}_\textrm{est}.\hat{\mathbf{l}}_\textrm{ps})$) as an error metric. Here, $\hat{\mathbf{l}}_\textrm{est}$ is a unit vector in the same direction as the mean of the posterior light estimate of \MTD and $\hat{\mathbf{l}}_\textrm{ps}$ is a unit vector in the same direction as the corresponding photometric stereo groundtruth. Again, these results indicate the superior performance of \MTD in comparison to other baselines in both real and synthetic image settings.

\begin{figure*}[h]
	\centering
	\setlength\fboxsep{0.2mm}
	\setlength\fboxrule{0pt}
	\begin{tikzpicture}
	
	\node at (-7.3, 2.9) {\small (a) Observed};
	\node at (-4.2, 2.9) {\small (b) Reflectance};
	\node at (-0.58, 2.9) {\small (c) Variance};
	\node at (2.35, 2.9) {\small (d) Light};
	\node at (6.4, 2.9) {\small (e) Normal};
	
	\node at (-4.15, 2.6) {\tiny $\color{gray}{\overbrace{\quad\quad\quad\quad\quad\quad\quad\quad\quad\quad
														  \quad\quad\quad\quad\quad\quad\quad\quad\quad\quad
														  \quad\quad\quad\quad\quad\quad}}$};
															  
    \node at (2.45, 2.6) {\tiny $\color{gray}{\overbrace{\quad\quad\quad\quad\quad\quad\quad\quad\quad\quad
														 \quad\quad\quad\quad\quad\quad\quad\quad\quad\quad}}$};
															  
	\node at (6.58, 2.6) {\tiny $\color{gray}{\overbrace{\quad\quad\quad\quad\quad\quad\quad\quad\quad\quad
														 \quad\quad\quad\quad\quad\,\,\,}}$};

		\matrix at (0, 0) [matrix of nodes, nodes={anchor=east}, column sep=-0.15cm, row sep=-0.2cm]
		{
			\fbox{\includegraphics[width=1cm]{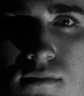}} & \hspace{0.12cm} & 
			\fbox{\includegraphics[width=1cm]{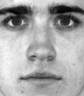}} &
			\fbox{\includegraphics[width=1cm]{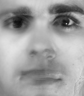}} &
			\fbox{\includegraphics[width=1cm]{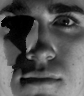}} &
			\fbox{\includegraphics[width=1cm]{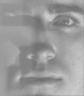}}  &
			\fbox{\includegraphics[width=1cm]{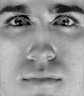}} & \hspace{0.12cm} & 
			\fbox{\includegraphics[width=1cm]{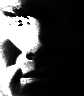}} & \hspace{0.12cm} & 
			\fbox{\includegraphics[width=1cm]{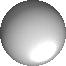}} &
			\fbox{\includegraphics[width=1cm]{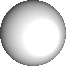}} &
			\fbox{\includegraphics[width=1cm]{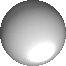}}  &
			\fbox{\includegraphics[width=1cm]{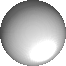}} & \hspace{0.12cm} & 
			\fbox{\includegraphics[width=1cm]{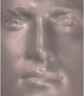}} &
			\fbox{\includegraphics[width=1cm]{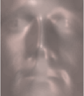}} &
			\fbox{\includegraphics[width=1cm]{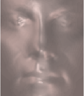}}  	\\
			
			\fbox{\includegraphics[width=1cm]{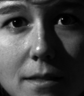}} & \hspace{0.12cm} & 
			\fbox{\includegraphics[width=1cm]{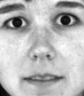}} &
			\fbox{\includegraphics[width=1cm]{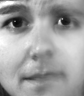}} &
			\fbox{\includegraphics[width=1cm]{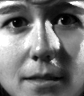}} &
			\fbox{\includegraphics[width=1cm]{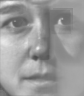}} &
			\fbox{\includegraphics[width=1cm]{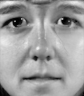}} & \hspace{0.12cm} & 
			\fbox{\includegraphics[width=1cm]{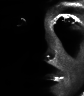}} & \hspace{0.12cm} & 
			\fbox{\includegraphics[width=1cm]{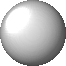}} &
			\fbox{\includegraphics[width=1cm]{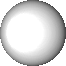}}  &
			\fbox{\includegraphics[width=1cm]{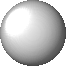}} &
			\fbox{\includegraphics[width=1cm]{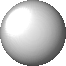}} & \hspace{0.12cm} & 
			\fbox{\includegraphics[width=1cm]{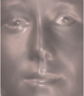}} &
			\fbox{\includegraphics[width=1cm]{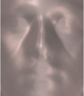}} &
			\fbox{\includegraphics[width=1cm]{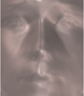}}  \\

			\fbox{\includegraphics[width=1cm]{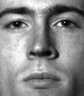}} & \hspace{0.12cm} & 
			\fbox{\includegraphics[width=1cm]{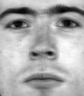}} &
			\fbox{\includegraphics[width=1cm]{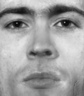}} &
			\fbox{\includegraphics[width=1cm]{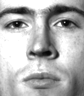}} &
			\fbox{\includegraphics[width=1cm]{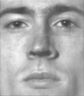}} &
			\fbox{\includegraphics[width=1cm]{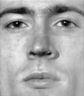}} & \hspace{0.12cm} & 
			\fbox{\includegraphics[width=1cm]{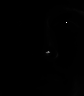}} & \hspace{0.12cm} & 
			\fbox{\includegraphics[width=1cm]{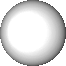}} &
			\fbox{\includegraphics[width=1cm]{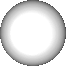}} &
			\fbox{\includegraphics[width=1cm]{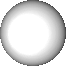}}  &
			\fbox{\includegraphics[width=1cm]{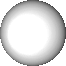}} & \hspace{0.12cm} & 
			\fbox{\includegraphics[width=1cm]{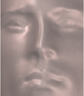}} &
			\fbox{\includegraphics[width=1cm]{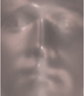}} &
			\fbox{\includegraphics[width=1cm]{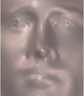}} \\
			
			\fbox{\includegraphics[width=1cm]{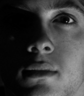}} & \hspace{0.12cm} & 
			\fbox{\includegraphics[width=1cm]{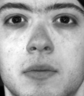}} &
			\fbox{\includegraphics[width=1cm]{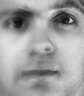}} &
			\fbox{\includegraphics[width=1cm]{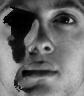}} &
			\fbox{\includegraphics[width=1cm]{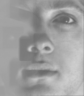}} &
			\fbox{\includegraphics[width=1cm]{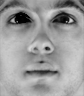}} & \hspace{0.12cm} & 
			\fbox{\includegraphics[width=1cm]{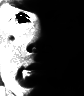}} & \hspace{0.12cm} & 
			\fbox{\includegraphics[width=1cm]{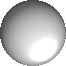}} &
			\fbox{\includegraphics[width=1cm]{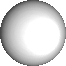}} &
			\fbox{\includegraphics[width=1cm]{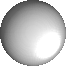}} &
			\fbox{\includegraphics[width=1cm]{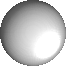}} & \hspace{0.12cm} & 
			\fbox{\includegraphics[width=1cm]{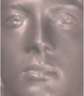}} &
			\fbox{\includegraphics[width=1cm]{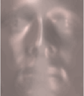}} &
			\fbox{\includegraphics[width=1cm]{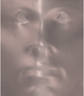}}  \\
	     };
	     
	     \node at (-6.4, -2.7) {\small GT};
	     \node at (-5.25, -2.7) {\small BU};
	     \node at (-4.1, -2.7) {\small MP};
	     \node at (-3.0, -2.7) {\small Forest};
	     \node at (-1.87, -2.7) {\small \textbf{CMP}};

	     \node at (-0.55, -2.7) {\small \textbf{CMP}};
	     
	     \node at (0.75, -2.7) {\small GT};
	     \node at (1.9, -2.7) {\small MP};
	     \node at (3, -2.7) {\small Forest};
	     \node at (4.15, -2.7) {\small \textbf{CMP}};
	     
	     \node at (5.5, -2.7) {\small GT};
	     \node at (6.6, -2.7) {\small MP};
	     \node at (7.72, -2.7) {\small \textbf{CMP}};

	\end{tikzpicture}
	\vspace{-0.4cm}
	\mycaption{A visual comparison of inference results}{For 4 randomly chosen test images, we show inference results obtained by competing methods. (a)~Observed images. (b)~Inferred reflectance maps. \textit{GT} is the photometric stereo groundtruth, \textit{BU} is the Biswas \etal (2009) reflectance estimate and \textit{Forest} is the consensus prediction. (c)~The variance of the inferred reflectance estimate produced by \MTD (normalized across rows). High variance regions correlate strongly with cast shadows. (d)~Visualization of inferred light directions. (e)~Inferred normal maps.}
	\label{fig:shading-qualitative-multiple-subjects}
\end{figure*}

\begin{figure*}[H]
	\centering
	\setlength\fboxsep{0.2mm}
	\setlength\fboxrule{0pt}
	\begin{tikzpicture}

		\matrix at (0, 0) [matrix of nodes, nodes={anchor=east}, column sep=-0.15cm, row sep=-0.2cm]
		{
			\fbox{\includegraphics[width=1cm]{shading/sample_3_1_X.png}} &
			\fbox{\includegraphics[width=1cm]{shading/sample_3_1_GT.png}} &
			\fbox{\includegraphics[width=1cm]{shading/sample_3_1_BISWAS.png}}  &
			\fbox{\includegraphics[width=1cm]{shading/sample_3_1_VMP.png}}  &
			\fbox{\includegraphics[width=1cm]{shading/sample_3_1_FOREST.png}}  &
			\fbox{\includegraphics[width=1cm]{shading/sample_3_1_CMP.png}}  &
			\fbox{\includegraphics[width=1cm]{shading/sample_3_1_CMPVAR.png}} & \hspace{0.25cm} & 
			
			\fbox{\includegraphics[width=1cm]{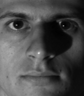}} &
			\fbox{\includegraphics[width=1cm]{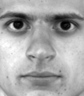}} &
			\fbox{\includegraphics[width=1cm]{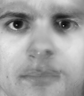}}  &
			\fbox{\includegraphics[width=1cm]{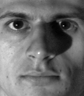}}  &
			\fbox{\includegraphics[width=1cm]{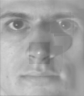}}  &
			\fbox{\includegraphics[width=1cm]{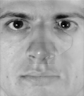}}  &
			\fbox{\includegraphics[width=1cm]{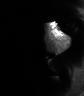}} 
			 \\
			
			\fbox{\includegraphics[width=1cm]{shading/sample_3_2_X.png}} &
			\fbox{\includegraphics[width=1cm]{shading/sample_3_2_GT.png}} &
			\fbox{\includegraphics[width=1cm]{shading/sample_3_2_BISWAS.png}}  &
			\fbox{\includegraphics[width=1cm]{shading/sample_3_2_VMP.png}}  &
			\fbox{\includegraphics[width=1cm]{shading/sample_3_2_FOREST.png}}  &
			\fbox{\includegraphics[width=1cm]{shading/sample_3_2_CMP.png}}  &
			\fbox{\includegraphics[width=1cm]{shading/sample_3_2_CMPVAR.png}} & \hspace{0.25cm} & 
			
			\fbox{\includegraphics[width=1cm]{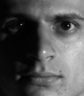}} &
			\fbox{\includegraphics[width=1cm]{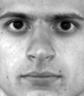}} &
			\fbox{\includegraphics[width=1cm]{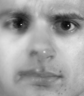}}  &
			\fbox{\includegraphics[width=1cm]{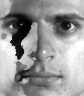}}  &
			\fbox{\includegraphics[width=1cm]{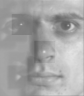}}  &
			\fbox{\includegraphics[width=1cm]{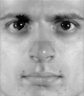}}  &
			\fbox{\includegraphics[width=1cm]{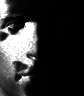}}
			 \\

			\fbox{\includegraphics[width=1cm]{shading/sample_3_3_X.png}} &
			\fbox{\includegraphics[width=1cm]{shading/sample_3_3_GT.png}} &
			\fbox{\includegraphics[width=1cm]{shading/sample_3_3_BISWAS.png}} &
			\fbox{\includegraphics[width=1cm]{shading/sample_3_3_VMP.png}}  &
			\fbox{\includegraphics[width=1cm]{shading/sample_3_3_FOREST.png}}  &
			\fbox{\includegraphics[width=1cm]{shading/sample_3_3_CMP.png}}  &
			\fbox{\includegraphics[width=1cm]{shading/sample_3_3_CMPVAR.png}} & \hspace{0.25cm} & 
			
			\fbox{\includegraphics[width=1cm]{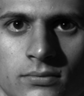}} &
			\fbox{\includegraphics[width=1cm]{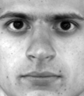}} &
			\fbox{\includegraphics[width=1cm]{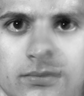}}  &
			\fbox{\includegraphics[width=1cm]{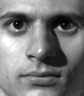}}  &
			\fbox{\includegraphics[width=1cm]{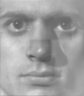}}  &
			\fbox{\includegraphics[width=1cm]{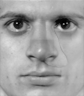}}  &
			\fbox{\includegraphics[width=1cm]{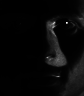}}
			 \\
	     };

	     \node at (-7.47, -2.0) {\small Observed};
	     \node at (-6.33, -2.0) {\small `GT'};
	     \node at (-5.2, -2.0) {\small BU};
	     \node at (-4.1, -2.0) {\small VMP};
	     \node at (-2.95, -2.0) {\small Forest};
	     \node at (-1.87, -2.0) {\small \textbf{CMP}};
	     \node at (-0.75, -2.0) {\small Variance};
	     
	     \node at (0.75, -2.0) {\small Observed};
	     \node at (1.87, -2.0) {\small `GT'};
	     \node at (2.95, -2.0) {\small BU};
	     \node at (4.1, -2.0) {\small VMP};
	     \node at (5.25, -2.0) {\small Forest};
	     \node at (6.35, -2.0) {\small \textbf{CMP}};
	     \node at (7.48, -2.0) {\small Variance};

	\end{tikzpicture}
	\mycaption{Robustness to varying illumination}{Reflectance estimation on two subject images with varying illumination. Left to right: observed image, photometric stereo estimate which is used as a proxy for groundtruth, bottom-up estimate of \cite{Biswas2009}, VMP result, consensus forest estimate, CMP mean, and CMP variance.}
	\label{fig:shading-qualitative-same-subject}
\end{figure*}

\begin{figure*}
	\centering
	\subfigure[Synthetic]{
		\includegraphics[width=0.46\linewidth]{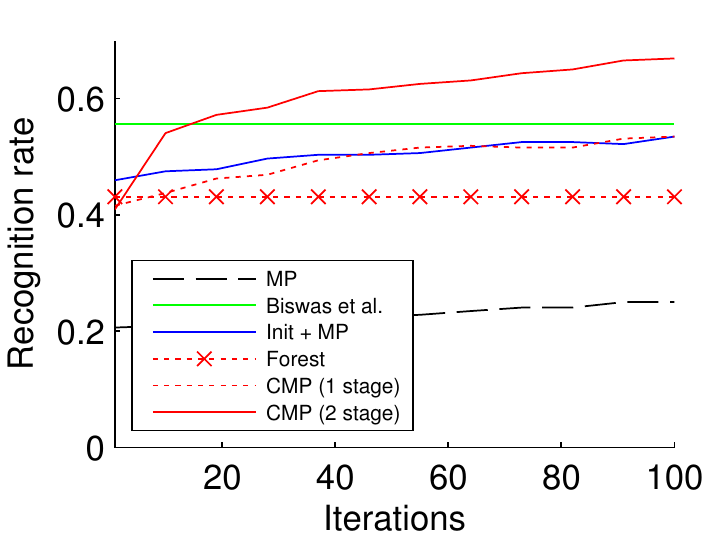}
	}
	\subfigure[Real]{
		\includegraphics[width=0.46\linewidth]{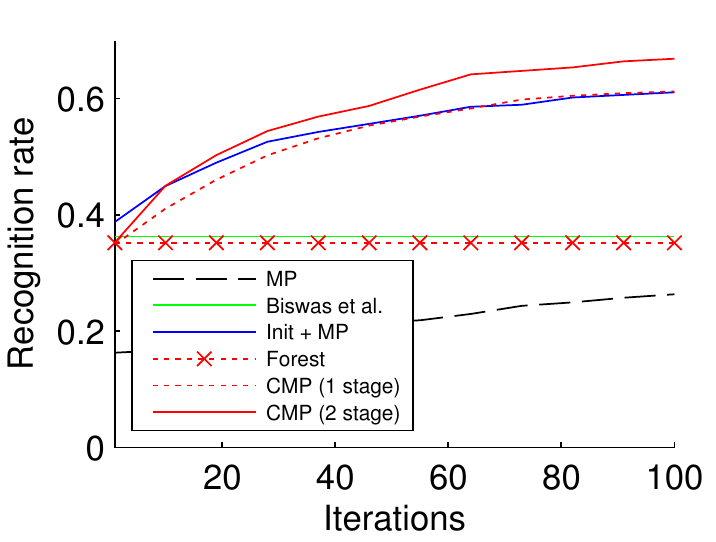}
	}
	\mycaption{Reflectance inference accuracy}{Results have been averaged over all images of test subjects. (a) Synthetic, shadowless images. (b) Real images.}
	\label{fig:shading-quantitative-reflectance}
\end{figure*}

\begin{figure*}
	\subfigure[Synthetic]{
		\includegraphics[width=0.46\linewidth]{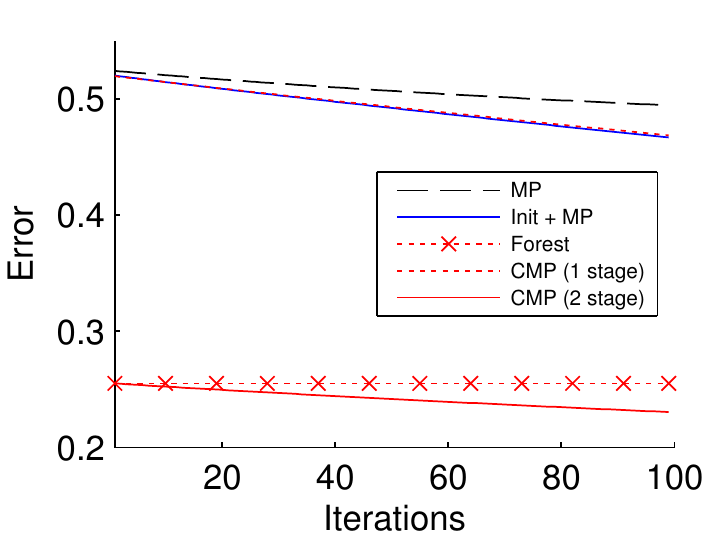}
	}
	\subfigure[Real]{
		\includegraphics[width=0.46\linewidth]{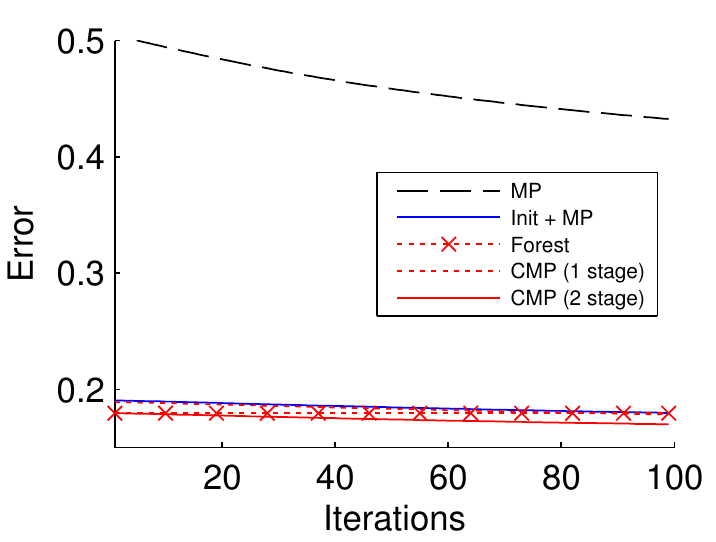}
	}
	\mycaption{Light inference accuracy}{Results have been averaged over all images of test subjects. (a) Synthetic, shadowless images. (b) Real images.}
	\label{fig:shading-quantitative-light}
\end{figure*}

\clearpage


\end{document}